\documentclass[11pt, a4paper, singlecolumn, confidential, copyright, goog]{google}

\usepackage[authoryear, sort&compress, round]{natbib}
\let\cite\citep
\usepackage{svg}
\usepackage{subcaption}
\usepackage{adjustbox}

\paperurl{https://github.com/google/GNM}

\uselogo{}
\usepackage{xcolor}
\usepackage{hyperref}
\usepackage{bm}
\title{GNM Head: A Generative aNthropometric Model of the human head}

\definecolor{orange}{RGB}{255,128,0}
\definecolor{lime}{RGB}{175,175,0}
\definecolor{darkblue}{RGB}{0,0,139}
\definecolor{crimson}{RGB}{153,0,0}
\definecolor{darkgreen}{RGB}{0,100,0}
\definecolor{deepslate}{RGB}{47,79,79}
\definecolor{midnight}{RGB}{25,25,112}

\correspondingauthor{gnm-owners@google.com,\newline *~Equal contribution \newline \textdagger~Equal contribution}

\author{Stylianos Ploumpis\textsuperscript{*}}
\author{Jan Bednarik\textsuperscript{*}}
\author{Gaspard Zoss}
\author{Ruslan Guseinov} 
\author{Luca Prasso}
\author{Prashanth Chandran} 
\author{Oliver Boyne}
\author{Vasileios Choutas}
\author{Timo Bolkart} 
\author{Daoye Wang}
\author{Menglei Chai}
\author{Di Qiu}
\author{Sebastian Winberg}
\author{Gilles Rainer}
\author{Lewis Bridgeman}
\author{Leonhard Helminger}
\author{Edo Collins}
\author{Delio Vicini}
\author{J\'er\'emy Riviere}
\author{Yannick Boetzel}
\author{Alexander Koumis}
\author{Stylianos Moschoglou}
\author{Jay Busch}
\author{Cynthia Herrera}
\author{Jacob Still}
\author{Scott Ysebert} 
\author{Peter Lincoln} 
\author{Sergio Orts Escolano}
\author{Christoph Rhemann}
\author{Erroll Wood\textsuperscript{\protect\textdagger}}
\author{Thabo Beeler\textsuperscript{\protect\textdagger}}
\author{Stefanos Zafeiriou\textsuperscript{\protect\textdagger}}

\affil{\thepa}
\begin{abstract}
Parametric models of the human head are essential tools traditionally used in computer vision and graphics for animation, rendering, and reconstruction. More recently, they serve as crucial conditioning signals within generative large vision models, allowing for tight spatial control of generated imagery. However, existing publicly available models are typically limited in anatomical scope, modeling only outer geometry while ignoring intra-oral and ocular structures, and frequently suffer from reduced geometric quality stemming from low-fidelity input datasets. In this report we introduce a new parametric model dubbed Generative aNthropometric Model (GNM), named as a homophone of the human genome. GNM encompasses the head, face, neck, eyeballs, teeth, and tongue, and it is built on an extensive database of high-resolution 3D scans combined with high-quality anatomy specific artist-made samples. This report details the data provenance, the model architecture including the specialized sub-models for the ocular and intra-oral structures, and shows its SotA performance on fitting target 3D face scans. To foster community innovation, the complete GNM framework is made publicly available.
\end{abstract}

\begin{document}

\newcommand{\real}{\mathbb{R}}
\newcommand{\integers}{\mathbb{Z}}
\newcommand{\lone}[1]{\left\lvert #1 \right\rvert}
\newcommand{\ltwo}[1]{\left\lVert #1 \right\rVert}
\newcommand{\frob}[1]{\lVert #1 \rVert_{\text{Frob}}}
\newcommand{\transpose}[1]{#1^{\top}}
\newcommand{\ones}{\mathbf{1}}
\newcommand{\zeros}{\mathbf{0}}
\newcommand{\eig}[1]{\operatorname{eig}\left(#1\right)}
\newcommand{\argmin}{\operatorname*{argmin}}

\newcommand{\verts}{V}
\newcommand{\gnmfunc}{\mathcal{M}}
\newcommand{\gnmparams}{\Theta}
\newcommand{\gnmdata}{\Psi}
\newcommand{\ngnmverts}{N_{\verts}}
\newcommand{\gnmvertdim}{3}
\newcommand{\paramsidsymbol}{\beta}
\newcommand{\paramsexprsymbol}{\phi}
\newcommand{\paramsrotsymbol}{\theta}
\newcommand{\paramstranslsymbol}{\tau}
\newcommand{\paramsid}{\bm{\paramsidsymbol}}
\newcommand{\paramsexpr}{\bm{\paramsexprsymbol}}
\newcommand{\paramsrot}{\bm{\paramsrotsymbol}}
\newcommand{\paramstransl}{\bm{\paramstranslsymbol}}
\newcommand{\paramsidi}[1]{\paramsidsymbol_{#1}}
\newcommand{\paramsexpri}[1]{\paramsexprsymbol_{#1}}
\newcommand{\paramsroti}[1]{\bm{\paramsrotsymbol}_{#1}}
\newcommand{\paramstransli}[1]{\paramstranslsymbol_{#1}}
\newcommand{\ngnmjoints}{K}
\newcommand{\nparamsid}{\lone{\paramsid}}
\newcommand{\nparamsexpr}{\lone{\paramsexpr}}
\newcommand{\nparamsrot}{\ngnmjoints \times 3}
\newcommand{\nparamstransl}{\gnmvertdim}
\newcommand{\gnmverttempl}{\mathbf{T}}
\newcommand{\gnmjointstempl}{\mathbf{J}}
\newcommand{\idbasis}{\mathbf{I}}
\newcommand{\exprbasis}{\mathbf{E}}
\newcommand{\jointsbasis}{\mathbf{Q}}
\newcommand{\skinningweights}{\mathbf{W}}
\newcommand{\jointparents}{\mathbf{p}}
\newcommand{\lbsfunc}{\mathcal{L}}
\newcommand{\vertsbind}{\mathbf{\verts_{B}}}
\newcommand{\skinningtf}{\mathbf{X}}
\newcommand{\blendweights}{\mathbf{W}}
\newcommand{\idexprfunc}{\mathcal{T}}
\newcommand{\skinningtffunc}{\mathcal{X}}
\newcommand{\jointsbind}{\mathbf{J}}
\newcommand{\jointsbindfunc}{\mathcal{J}}
\newcommand{\homogfunc}{\mathcal{H}}
\newcommand{\homogfuncinv}{\homogfunc^{-1}}
\newcommand{\vertex}{\mathbf{v}}
\newcommand{\nparamsidregion}[1]{\lone{\paramsid^{(\text{#1})}}}
\newcommand{\nparamsexprregion}[1]{\lone{\paramsexpr^{(\text{#1})}}}

\newcommand{\headregion}{head}
\newcommand{\idbasishead}{\mathbf{I^{(\text{\headregion})}}}
\newcommand{\datasetneutral}{\mathbf{X_{N}}}
\newcommand{\fulldatasetneutral}{\mathbf{\widehat{\datasetneutral}}}
\newcommand{\centereddatasetneutral}{\overline{\datasetneutral}}
\newcommand{\datasetneutralmean}{\overline{\mathbf{x_{N}}}}
\newcommand{\nsamplesdatasetneutral}{N_{N}}
\newcommand{\idbasisvectors}{\overline{\idbasis}}
\newcommand{\nidbasisvectors}{N_{I}}
\newcommand{\idbasiseigvalssymbol}{\Lambda}
\newcommand{\idbasiseigvals}{\bm{\idbasiseigvalssymbol}}
\newcommand{\idbasiseigvalsi}[1]{\idbasiseigvalssymbol_{#1}}
\newcommand{\covmatid}{\mathbf{C_{I}}}
\newcommand{\pcacall}[1]{\text{PCA}\left( #1 \right)}
\newcommand{\flattenfunc}{\mathcal{F}}
\newcommand{\flattenfuncinv}{\flattenfunc^{-1}}
\newcommand{\jointregressor}{\mathfrak{R}}
\newcommand{\headskinidbasis}{\mathbf{I^{(\text{\headregion})}_{s}}}
\newcommand{\headskinidbasisi}[1]{\mathbf{I^{(\text{\headregion})}_{s_{#1}}}}
\newcommand{\paramsidhead}{\paramsid^{(\text{\headregion})}}
\newcommand{\paramsidheadt}{\paramsid^{(\text{\headregion})^{\top}}}
\newcommand{\nparamsidhead}{\lone{\paramsidhead}}

\newcommand{\lefteyeregion}{left eye}
\newcommand{\righteyeregion}{right eye}
\newcommand{\lowerfaceregion}{lower face}
\newcommand{\tongueregion}{tongue}
\newcommand{\pupilregion}{pupil}
\newcommand{\vertssubjneut}[1]{\mathbf{\verts}_{\mathbf{n}}^{(#1)}}
\newcommand{\vertssubjexpr}[2]{\mathbf{\verts}_{\mathbf{e}_{#2}}^{(#1)}}
\newcommand{\exprdelta}[2]{\bm{\delta}^{(#1)}_{#2}}
\newcommand{\datasetexpressive}{\mathbf{X_{E}}}
\newcommand{\datasetexpressiveregion}[1]{\mathbf{X_{E}}^{(#1)}}
\newcommand{\nsamplesdatasetexpressive}{N_{M}}
\newcommand{\nsubjects}{N_{s}}
\newcommand{\nsubjectsamples}[1]{n_{#1}}
\newcommand{\vertexmaskregion}[1]{S_{#1}}
\newcommand{\exprbasisvectors}{\overline{\exprbasis}}
\newcommand{\exprbasisvectorsregion}[1]{\exprbasisvectors^{(\text{#1})}}
\newcommand{\exprbasisvectorsregioni}[2]{\exprbasisvectors_{#1}^{(\text{#2})}}
\newcommand{\exprbasiseigvalssymbol}{\Pi}
\newcommand{\exprbasiseigvals}{\bm{\exprbasiseigvalssymbol}}
\newcommand{\exprbasiseigvalsregion}[1]{\exprbasiseigvals^{(\text{#1})}}
\newcommand{\exprbasiseigvalsregioni}[2]{\exprbasiseigvalssymbol_{#1}^{(\text{#2})}}
\newcommand{\upcacall}[1]{\text{uPCA}\left( #1 \right)}
\newcommand{\covmatexpr}{\mathbf{C_{E}}}
\newcommand{\covmatexprregion}[1]{\covmatexpr^{(\text{#1})}}
\newcommand{\nexprbasisvectors}{N_{E}}
\newcommand{\nexprbasisvectorsregion}[1]{N_{\text{#1}}}
\newcommand{\exprbasisregion}[1]{\mathbf{E^{(\text{#1})}}}
\newcommand{\exprbasisregioni}[2]{\mathbf{E_{#1}^{(\text{#2})}}}
\newcommand{\exprbasislefteye}{\mathbf{E^{(\text{\lefteyeregion})}}}
\newcommand{\exprbasisrighteye}{\mathbf{E^{(\text{\righteyeregion})}}}
\newcommand{\exprbasislowerface}{\mathbf{E^{(\text{\lowerfaceregion})}}}
\newcommand{\exprbasistongue}{\mathbf{E^{(\text{\tongueregion})}}}
\newcommand{\exprbasispupil}{\mathbf{E^{(\text{\pupilregion})}}}
\newcommand{\paramsexprlefteye}{\paramsexpr^{(\text{\lefteyeregion})}}
\newcommand{\paramsexprlefteyet}{\paramsexpr^{(\text{\lefteyeregion})^{\top}}}
\newcommand{\nparamsexprlefteye}{\lone{\paramsexprlefteye}}
\newcommand{\paramsexprrighteye}{\paramsexpr^{(\text{\righteyeregion})}}
\newcommand{\paramsexprrighteyet}{\paramsexpr^{(\text{\righteyeregion})^{\top}}}
\newcommand{\nparamsexprrighteye}{\lone{\paramsexprrighteye}}
\newcommand{\paramsexprlowerface}{\paramsexpr^{(\text{\lowerfaceregion})}}
\newcommand{\paramsexprlowerfacet}{\paramsexpr^{(\text{\lowerfaceregion})^{\top}}}
\newcommand{\nparamsexprlowerface}{\lone{\paramsexprlowerface}}
\newcommand{\paramsexprtongue}{\paramsexpr^{(\text{\tongueregion})}}
\newcommand{\paramsexprtonguet}{\paramsexpr^{(\text{\tongueregion})^{\top}}}
\newcommand{\nparamsexprtongue}{\lone{\paramsexprtongue}}
\newcommand{\paramsexprpupil}{\paramsexpr^{(\text{\pupilregion})}}
\newcommand{\paramsexprpupilt}{\paramsexpr^{(\text{\pupilregion})^{\top}}}
\newcommand{\nparamsexprpupil}{\lone{\paramsexprpupil}}

\newcommand{\teethregion}{teeth}
\newcommand{\idbasisteeth}{\mathbf{I^{(\text{\teethregion})}}}
\newcommand{\paramsidteeth}{\paramsid^{(\text{\teethregion})}}
\newcommand{\paramsidteetht}{\paramsid^{(\text{\teethregion})^{\top}}}
\newcommand{\nparamsidteeth}{\lone{\paramsidteeth}}
\newcommand{\datasetteeth}{\mathbf{X_{T}}}
\newcommand{\datasetteethmean}{\overline{x_{T}}}
\newcommand{\nsamplesdatasetteeth}{N_{T}}

\newcommand{\eyeballregion}{eye}
\newcommand{\idbasiseyeballs}{\mathbf{I^{(\text{\eyeballregion})}}}
\newcommand{\paramsideyeballs}{\paramsid^{(\text{\eyeballregion})}}
\newcommand{\paramsideyeballst}{\paramsid^{(\text{\eyeballregion})^{\top}}}
\newcommand{\nparamsideyeballs}{\lone{\paramsideyeballs}}

\newcommand{\datasettongue}{\mathbf{X_{G}}}
\newcommand{\datasettonguemean}{\overline{\mathbf{x_{G}}}}
\newcommand{\nsamplesdatasettongue}{N_{G}}

\newcommand{\nseqframes}{F}
\newcommand{\nviews}{U}
\newcommand{\seqframe}{f}
\newcommand{\view}{u}
\newcommand{\seqheadverts}{\mathbf{V}}
\newcommand{\landmarks}{\mathbf{l}}
\newcommand{\tensorgnmfunc}{\dot{\gnmfunc}}
\newcommand{\confweight}{\omega}
\newcommand{\landmarkextractor}{\mathcal{A}}
\newcommand{\nlandmarks}{L}

\maketitle
\section{Introduction}
\label{sec:intro}

The digital representation of human appearance remains a principal driver of research at the intersection of computer vision and graphics, demanding frameworks that are compact, expressive, and controllable. Addressing this demand, 3D Morphable Models (3DMMs) have emerged as foundational statistical frameworks that distill the complex 3D geometry and appearance of the human head into a low-dimensional, controllable latent space \cite{blanz1999morphable}. Today, these models have evolved into essential priors for foundational large vision systems. In 2D generative AI, 3DMM parameters act as geometric anchors that condition diffusion models \cite{zhang2023adding, chen2024morphable}, resolving the spatial and temporal instabilities inherent in unguided generation \cite{rombach2022high}. Concurrently, they enable the generation of millions of diverse, privacy-compliant synthetic humans \cite{gdpr2016, wood2021fakeit, varol2017learning}, grounding AI training in physical plausibility. Furthermore, 3DMMs serve as a crucial structural bridge for state-of-the-art neural rendering techniques such as 3D Gaussian Splatting (3DGS) \cite{kerbl20233d} and Neural Radiance Fields (NeRFs) \cite{mildenhall2021nerf}. By anchoring neural primitives to a physiologically accurate parametric surface \cite{qian2024gaussianavatars, xu2024gaussian}, hybrid frameworks can achieve the photorealism of neural rendering whilst mitigating non-physical deformations, enabling expressive, real-time animation of complex facial features \cite{peng2026parametric, giebenhain2024npga}.

Beyond foundational generation and rendering, 3DMMs are indispensable across a broad spectrum of applied research. They are the de facto standard for speech-driven audio-visual synthesis, driving the realistic lip synchronization and coarticulation vital for telepresence \cite{cudeiro2019capture, fan2022faceformer, aneja2024facetalk, sun2024diffposetalk, danecek2025supervising}. In the entertainment sector, these parametrized models form the backbone of high-fidelity performance capture for cinematic visual effects and video games \cite{edwards2020jali, alexander2010digital, beeler2011high, epicgames2026metahuman}. Within media forensics, they provide essential geometric constraints—such as 3D pose and landmark consistencies—to bypass the spatial limitations of 2D detectors and accurately identify AI-manipulated deepfakes \cite{peng2025within, petmezas2025video}. Finally, their rigorous mathematical parameterization extends into medicine, facilitating craniofacial surgery planning, syndrome classification, and quantitative biometric shape analysis \cite{o2022growth, o20213d, Egger2020:3DMM}.

\begin{figure}
    \centering
    \includegraphics[width=1.0\textwidth]{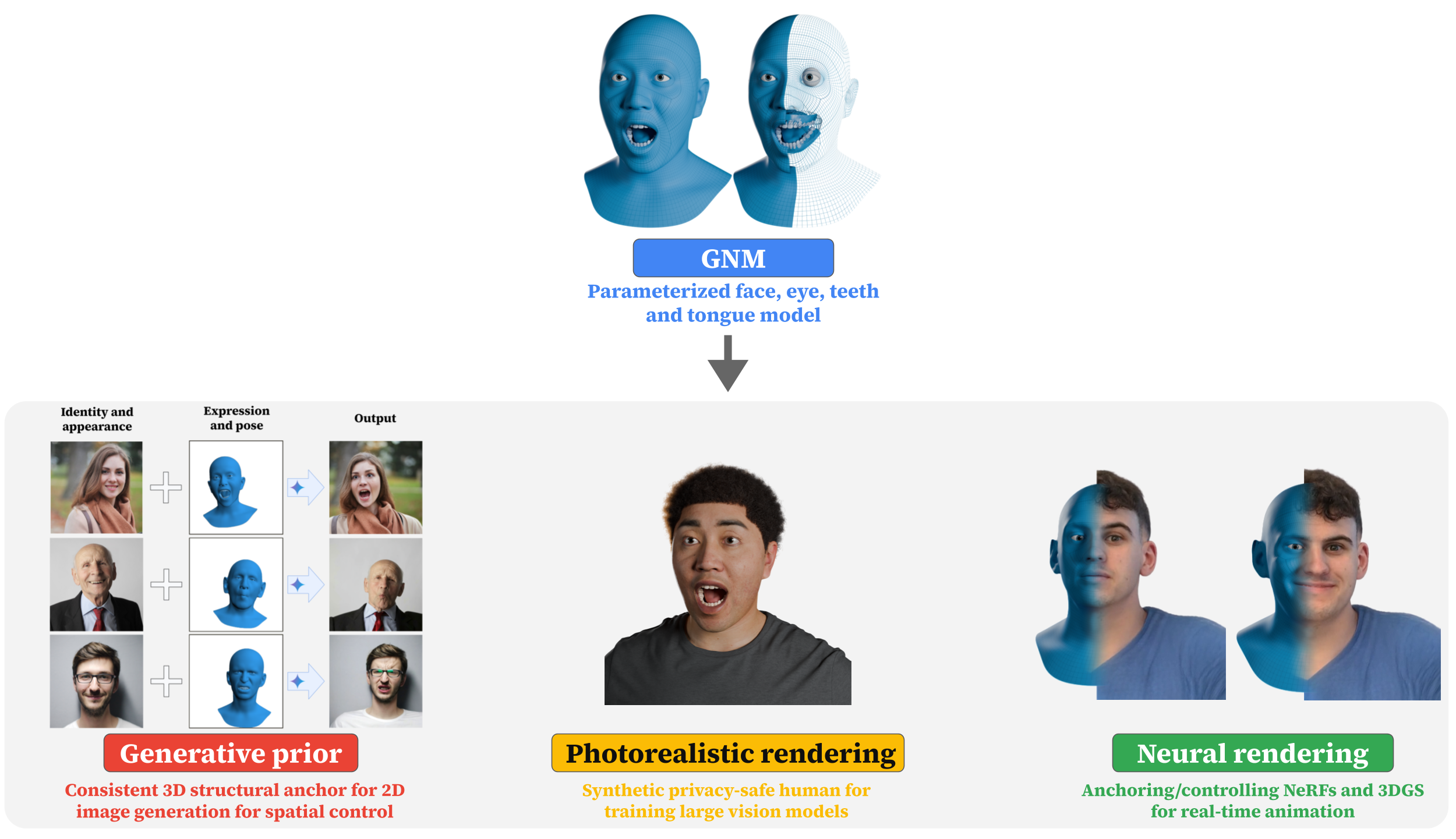}
    \caption{Overview of GNM and its applications. GNM is a holistic parametric model encompassing the face, eyes, teeth, and tongue within a unified statistical space. This complete anatomical representation serves as a robust 3D prior for multiple downstream modalities. These include enabling structurally consistent 2D generative modeling, generating privacy-safe photorealistic datasets for computer vision, and powering real-time neural rendering applications such as 3D Gaussian Splatting.}
    \label{fig:splash}
\end{figure}

Despite their widespread use, existing parametric head models, such as FLAME \cite{li2017learning}, Basel Face Model (BFM) \cite{paysan20093d} or Large Scale Face Model (LSFM) \cite{booth20163d}, suffer from a fundamental limitation: they treat the human head as a hollow shell, almost entirely omitting internal oral anatomy such as the teeth and tongue, and fine ocular structures. This structural omission limits the downstream applications. Without these internal assets, generative models and neural pipelines lack the necessary geometric constraints to synthesize realistic mouth interiors, leading to a substantial drop in visual quality of the generated facial performances. Furthermore, frameworks lose fine-grained semantic control over critical non-verbal communication cues, such as precise lip or tongue co-articulation.

Driven by the growing demand for physical plausibility in AI and the need to overcome the "Uncanny Valley" in immersive Augmented Reality (AR) and telepresence, we address these limitations by introducing GNM, a holistic parametric head framework that unifies the external facial skin, eyes, teeth, and tongue within a single statistical space. In contrast to current vanilla 3DMMs, GNM introduces three key advancements. First, GNM integrates teeth geometry and embeds their structural shape variation directly within the global identity shape space. Second, our framework incorporates explicit tongue blendshapes and localised facial expression blendshapes, moving beyond standard global expressions to allow for fine-grained expression control. Finally, GNM models pupil dilation as well as sclera and cornea shape varying with human identity.
The model is built on a high-resolution mesh topology and trained on a large scale high-fidelity 3D facial dataset combined with artist-made specialised assets, yielding high reconstruction accuracy. It can serve as a robust 3D prior across multiple downstream modalities, such as structurally consistent 2D generative modelling, privacy-safe photorealistic dataset generation, and real-time neural rendering applications such as 3D Gaussian Splatting (see Figure~\ref{fig:splash}).

To foster innovation and lower the barrier to entry for high-fidelity digital humans, the GNM model is made publicly available to the global community, licensed for both academic research and commercial applications. To enable intuitive control, we develop a Semantic Sampler using a dual-CVAE architecture that maps high-level demographic and expression attributes on to a smooth parametric manifold without unnatural geometric distortions. Additionally, we present a fitting pipeline featuring specialized collision constraints, localized tongue convex hull tests, and regularizers, to reconstruct GNM meshes from single-view or multi-view images based on dense 2D facial landmarks. Extensive quantitative and qualitative evaluations demonstrate that GNM consistently outperforms FLAME \cite{li2017learning} across these downstream modalities, establishing its practical superiority in both geometric tracking fidelity and generative expressiveness.

\section{Related Models}
\label{sec:related_models}

The trajectory of 3DMMs began with the seminal work of \citet{blanz1999morphable}, who utilized Principal Component Analysis (PCA) to represent facial shape and texture as linear combinations of exemplar meshes. This foundation paved the way for widely adopted models like the BFM \cite{paysan20093d} and FaceWarehouse \cite{cao2013facewarehouse}, which established early standards for identity and expression variation. More recently, the FLAME model \cite{li2017learning} introduced a more anatomically flexible head model by incorporating skeletal articulation for the neck, jaw, and eyeballs. However, these traditional models focus primarily on the external facial mask, frequently neglecting the internal structures, such as the teeth and tongue. To capture a wider demographic variance, the LSFM \cite{booth20163d} leveraged 10,000 diverse facial scans to create a highly robust statistical foundation. This extensive dataset subsequently served as the foundation for constructing a Universal Head Model (UHM) \cite{ploumpis2019combining, ploumpis2020towards}, which successfully expanded this statistical footprint to encompass the full cranium, scalp, and ears. While these universal head variants expand the statistical shape space to the full cranium, traditional global PCA formulations introduce long-range coupled deformations, where altering a facial parameter might unintentionally deform the skull structure. In contrast, GNM utilizes localized part-based smoothness constraints over the cranium, scalp, and the back of the ears, providing a highly decoupled, semantically isolated representation with greater regional statistical variance.

Contemporary state-of-the-art advancements have sought to overcome the lack of high-frequency mesh detail through sophisticated capture frameworks, such as DECA (Detailed Expression Capture and Animation) \cite{feng2021learning} and EMOCA (Emotion-driven Monocular Face Capture) \cite{danecek2022emoca}. While DECA utilizes regressed displacement maps to capture fine-scale wrinkles and EMOCA enhances the capture of emotional nuances, these methods still largely rely on the underlying FLAME topology. Consequently, they inherit its architectural drawbacks, including a simplified oral cavity and a lack of integrated, physiologically accurate eye models, which restricts their utility in high-end animation where internal mouth visibility is high and accurate representation of eyeballs is desired.

\begin{figure}[t]
       \centering
    \includegraphics[width=\textwidth]{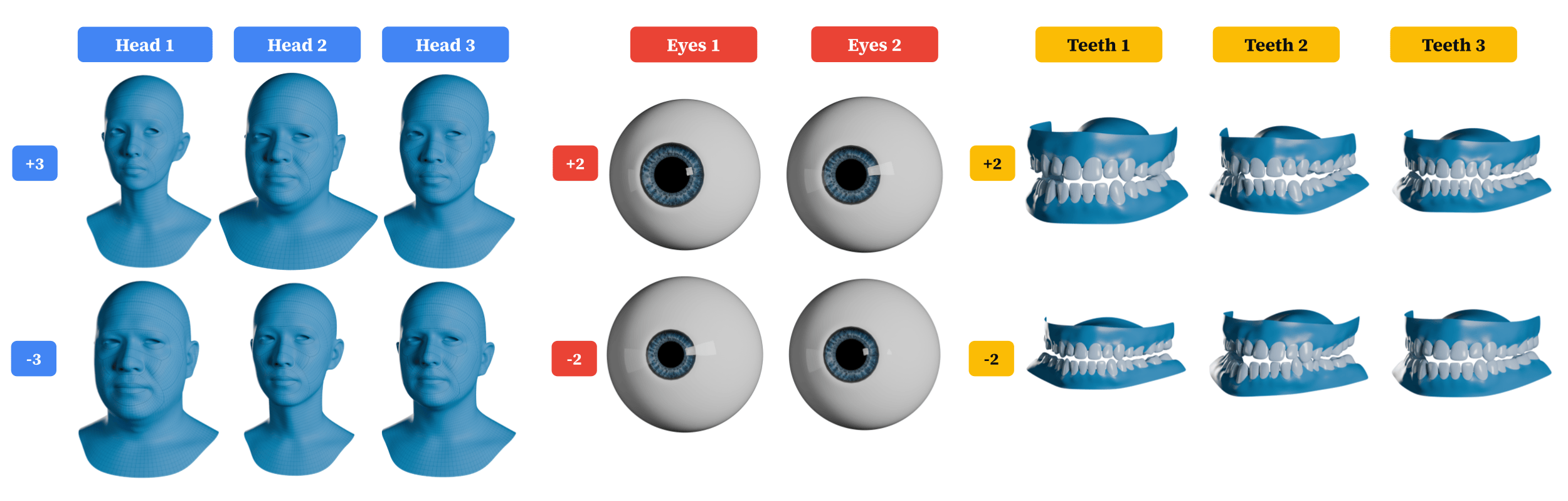}
    \includegraphics[width=\textwidth]{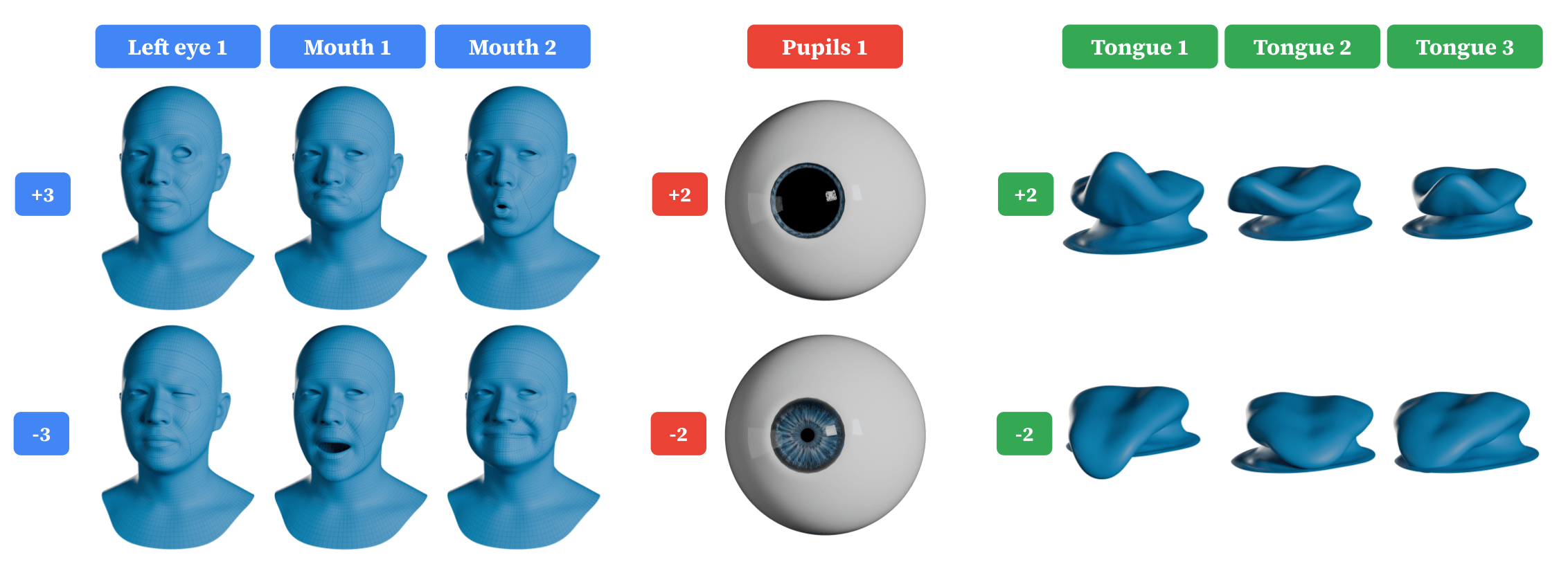}
    \caption{Principal components of the GNM statistical model. We illustrate the part-based formulation of the GNM manifold. The top panel displays the identity basis controlling structural variations of the skin, exterior eyeballs, and teeth. The bottom panel shows the expression basis, which governs the dynamic deformations of the regional skin components, internal pupil dilation, and tongue poses.}
    \label{fig:gnm_bases}
\end{figure}

To bypass the resolution boundaries of explicit meshes, neural parametric head models, such as NPHM \cite{giebenhain2023learning}, imHead \cite{potamias2025imhead}, the Shape Transformer \cite{chandran2022shape} or AIM \cite{chandran2024anatomically} refrain from utilizing explicit linear bases in favor of implicit neural representations. Others also explored linking these neural models with physical simulation \cite{srinivasan2021learning, yang2022implicit, yang2023implicit, yang2024learning}. While these models gracefully handle topological changes like mouth opening and close-up details, their `black-box' latent spaces lack the intuitive, semantically disentangled control required by standard computer graphics pipelines. This makes it challenging for an animator to target specific operations such as dental alignment or pupil dilation without altering adjacent facial regions.

The most recent frameworks of neural frameworks, including StyleMorpheus \cite{yan2025stylemorpheus} and Gaussian Head Avatar \cite{xu2024gaussian, chu2024generalizable}, leverage Neural Radiance Fields (NeRFs) \cite{mildenhall2021nerf} or 3DGS \cite{kerbl20233d} to achieve photorealistic rendering of complex features like hair and accessories. While these methods produce impressive visual results, they are often identity specific, require significant compute for real-time animation or tie the geometric primitives to the underlying 3DMM parameters or geometry \cite{buehler2024cafca,qian2024gaussianavatars} suggesting the strong need for the geometry prior. Furthermore, they frequently `bake' the teeth and eyes into the neural volume, making it difficult to achieve the precise, coordinated movement between the lips and tongue necessary for realistic speech. Rare exceptions attempt to build general neural priors such as SynShot \cite{zielonka2025synthetic} or GPHM \cite{xu2025gphm} yet they remain bound to the visual layer and cannot guarantee the coordinated physical boundaries between the lips, teeth, and tongue required for dynamic speech. To address this, specialized part-specific submodels have been developed to isolate individual intraoral features \cite{medina2022speech, ploumpis20223d} however, they exist as standalone assets lacking a unified, multi-part registration manifold.

Recognizing the limitations of global facial topologies in capturing complex internal mechanics, a parallel line of research has focused on modeling highly specialized, region-specific anatomical components. For the ocular region, classical parametric models have been developed to accurately represent the intricate geometry and refractive properties of the sclera, cornea, and iris \cite{berard2014high, berard2016lightweight, berard2019practical, wood20163d}, while recent non-linear approaches like EyeNeRF \cite{li2022eyenerf} leverage volumetric rendering for gaze-dependent photorealism. Similarly, the complex articulating mechanics of the dental arches have been addressed through robust linear and non-linear teeth models \cite{abdelrehim20132d, wu2016model, zhang2022implicit} and biomechanical jaw rigs \cite{zoss2018empirical, zoss2019accurate, yang2019jaw}. The highly deformable tongue, which is critical for accurate phonetic articulation, has been explored through both linear blendshape formulations \cite{medina2022speech,ploumpis20223d} and modern neural paradigms designed to handle extreme intraoral expressions \cite{giebenhain2025joker}. Other regional works have isolated the ears \cite{dai2018data, zhou2017deformable} to build distinct morphable models capable of capturing fine cartilaginous variations. While these part-based representations achieve unprecedented regional fidelity spanning both classical PCA and modern deep-learning architectures they overwhelmingly exist as standalone, isolated assets. Integrating these disparate regional priors into a unified, mathematically consistent manifold remains a fundamental challenge.

GNM aims to bridge the gap between these approaches by combining the simplicity and controllability of a linear parametric model with the holistic anatomical scope of neural avatars. By leveraging a large-scale 3D facial database, GNM incorporates dedicated components for the teeth, tongue, and eyes as shown in the Figure~\ref{fig:gnm_bases}. This ensures that the model remains compatible with standard graphics workflows while providing the high-fidelity detail particularly in the perioral and ocular regions that current SotA models still struggle to represent in a unified, controllable manifold.
\section{GNM}
\label{sec:GNM}

GNM is a statistical model comprising a linear identity and expression basis, a skeletal rig for neck and eyeballs articulation, joint location identity basis and standard linear blend skinning (LBS) for posing the mesh vertices. The model conceptually follows a standard 3DMM definition \cite{blanz1999morphable, li2017learning} with a few notable improvements: (i) the model is built on a combination of a large high-quality real-world dataset of registered expressive human faces and an artist-created synthetic one; (ii) the model contains a diverse subspace of human teeth shapes and tongue poses; and (iii) the expression subspace is split into separate facial regions for higher-granularity control. The GNM owes its fidelity to the high-quality dataset, detailed in Section~\ref{sec:datafoundation}, obtained using a custom face registration pipeline described in section~\ref{sec:face_registration}. We formally define GNM in Section~\ref{sec:gnm_formulation}.

\subsection{Data Foundation \& Acquisition}
\label{sec:datafoundation}

\begin{figure}[ht]
    \centering
    \resizebox{\textwidth}{!}{%
        \includegraphics[height=5cm]{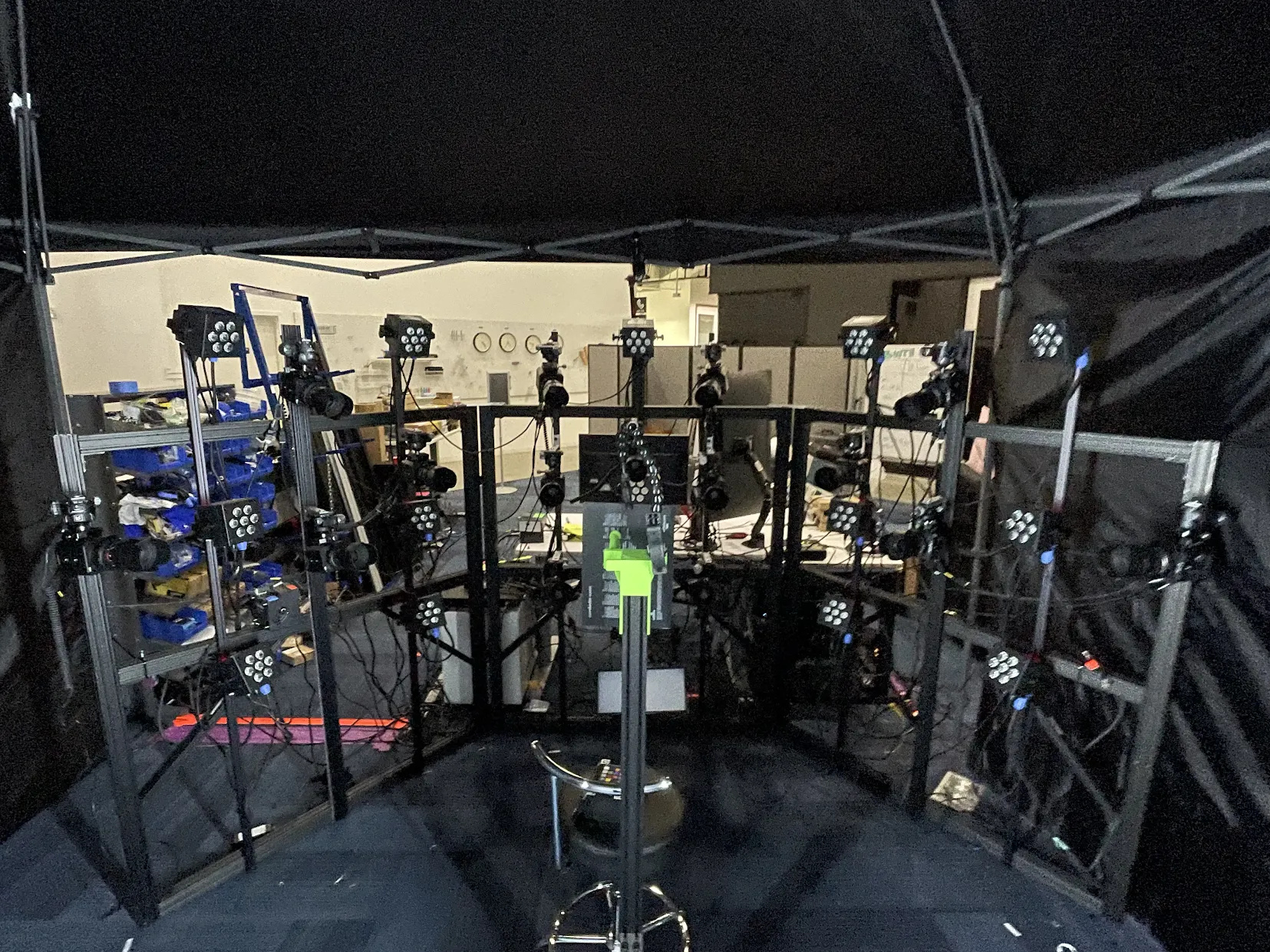}%
        \includegraphics[height=5cm]{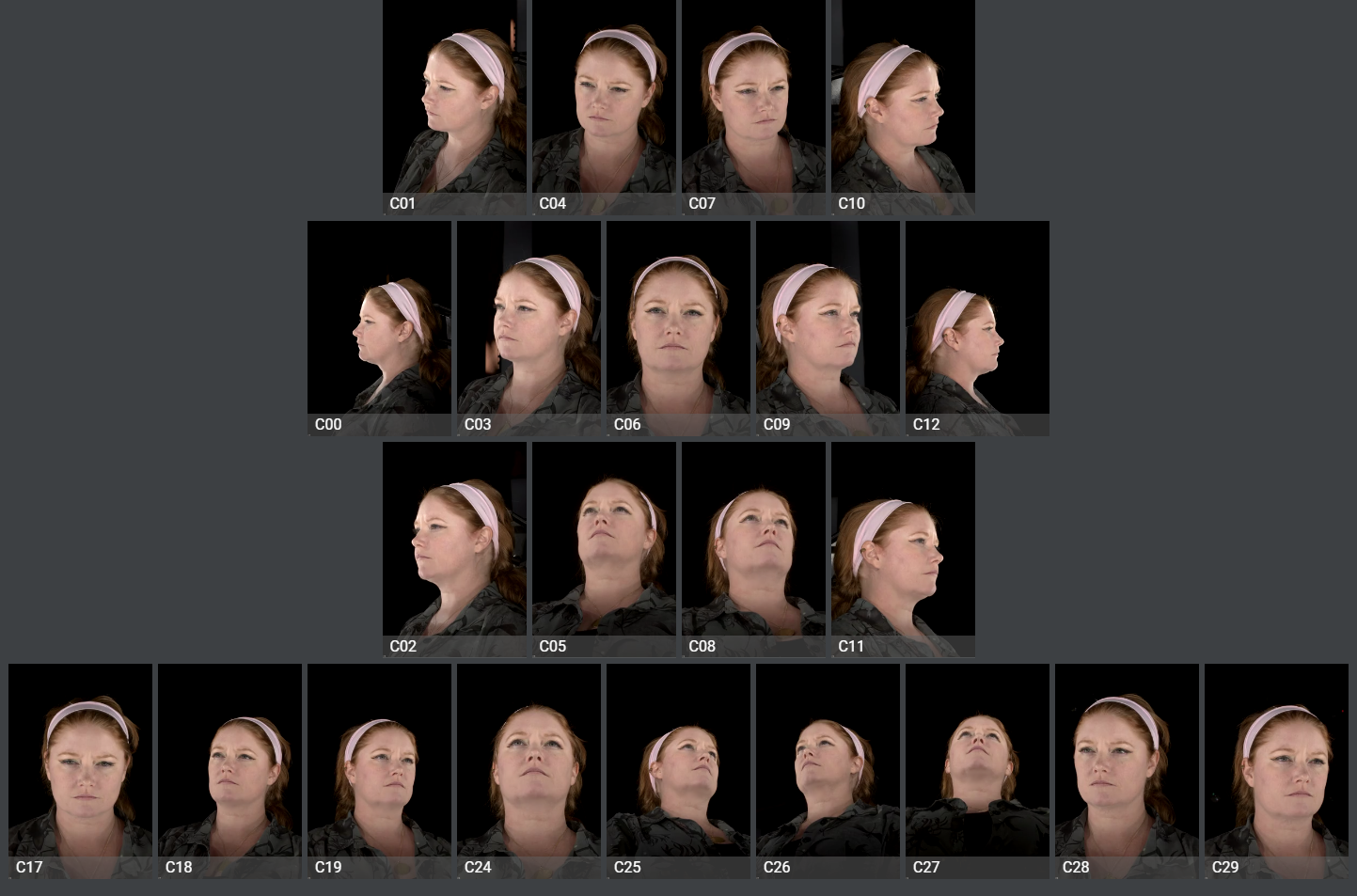}%
        \includegraphics[height=5cm]{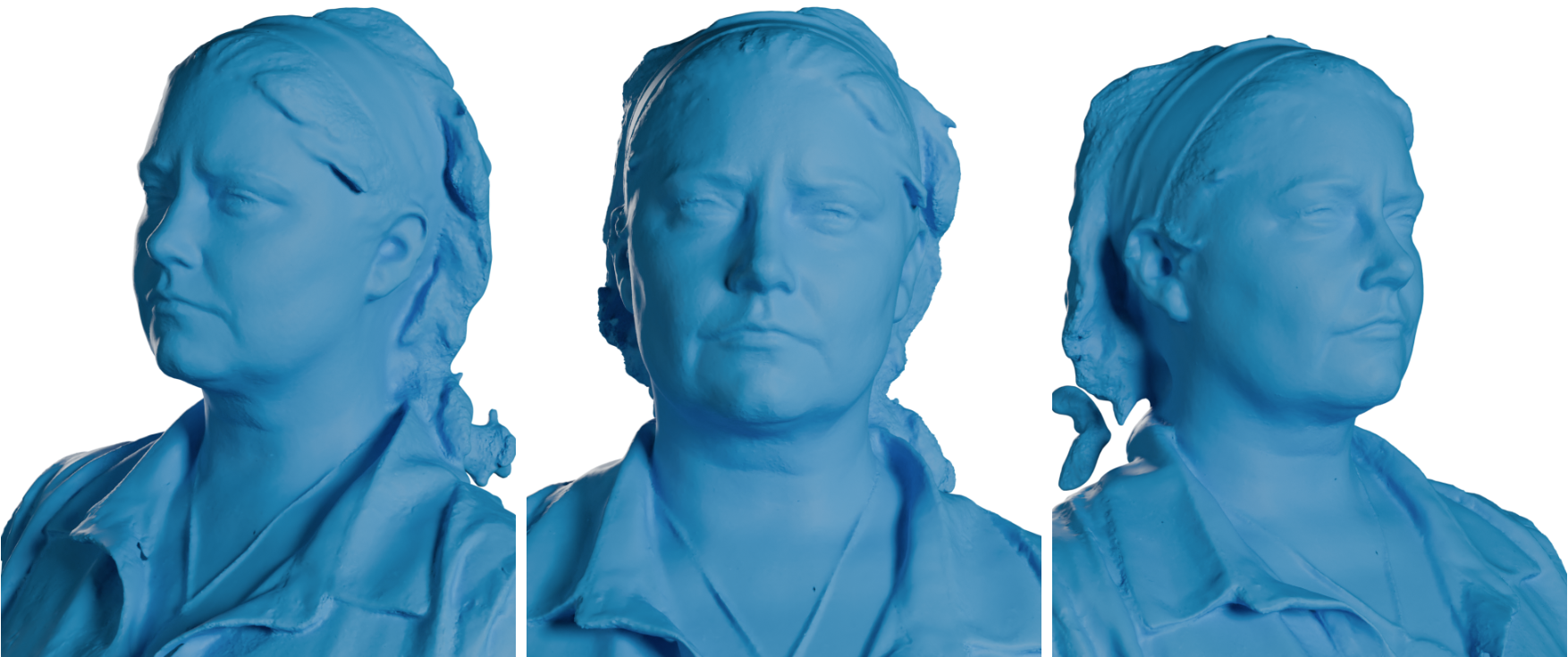}
    }
    \caption{Custom Multi-View Capture System: (Left) The physical acquisition rig featuring 22 high-resolution cameras and 14 controllable lights designed for uniform, diffused illumination. (Middle) Synchronized multi-view image streams capturing the subject across a 150-degree horizontal and 60-degree vertical span. (Right) The resulting high-fidelity 3D raw geometry produced by our multi-view depth refinement pipeline.}
    \label{fig:hb_system}
\end{figure}

The empirical foundation of the GNM framework is a large-scale, high-resolution 3D facial database meticulously curated to maximize morphological and demographic diversity. Raw geometry is acquired via a custom multi-view capture system equipped with synchronized cameras \cite{beeler2010high, guo2019relightables}. Our custom multi-view capture (Figure~\ref{fig:hb_system}) system features 22 high-resolution (6144 x 4096) ZCam E2 S6G cameras and 14 controllable lights, programmed to provide uniform, diffused illumination to maximize data quality and  subject comfort. Our capture setup is designed to span roughly 150 degrees horizontally and 60 degrees vertically in front of the subject in order to reconstruct the subject’s face at high fidelity using a multi-view depth refinement pipeline \cite{qiu2025chosen}. The capture protocol comprises an acquisition of a neutral relaxed expression and a set of static facial expressions.

\begin{figure}[ht]
    \centering
    \includegraphics[width=\textwidth]{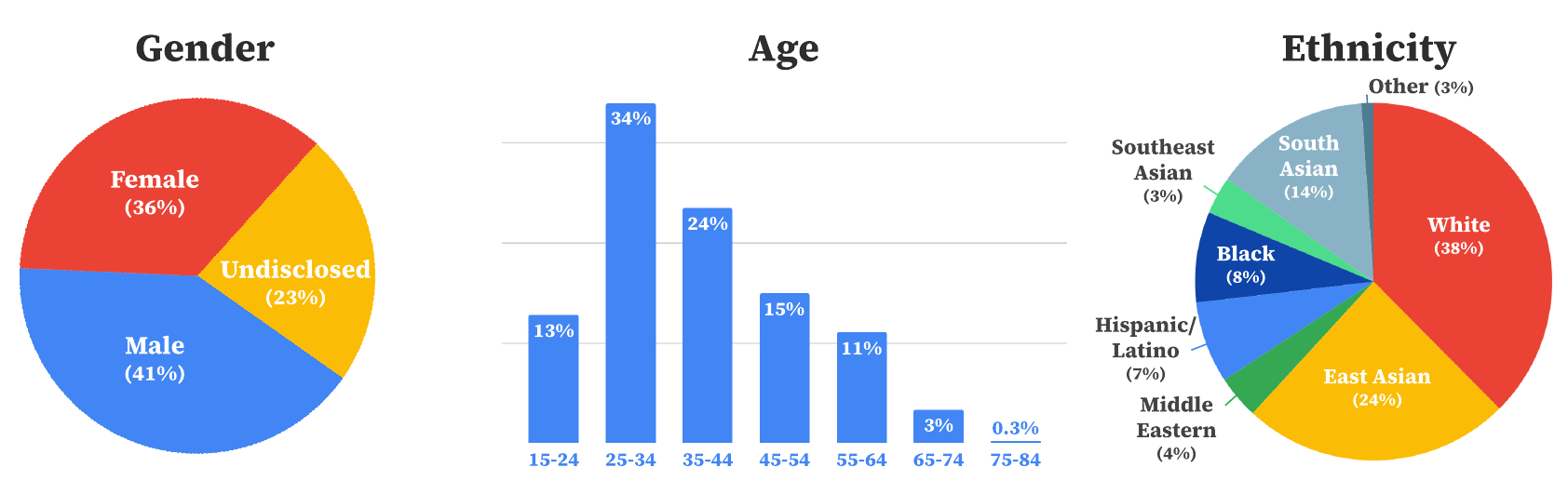}
    \caption{Distribution of the captured subjects across \emph{gender}, \emph{age}, and \emph{ethnicity}, highlighting the multi-demographic diversity required to construct a highly generalizable 3D morphable model.}
    \label{fig:hb_stats}
\end{figure}

\begin{figure}[ht]
    \centering    
    \includegraphics[width=\textwidth]{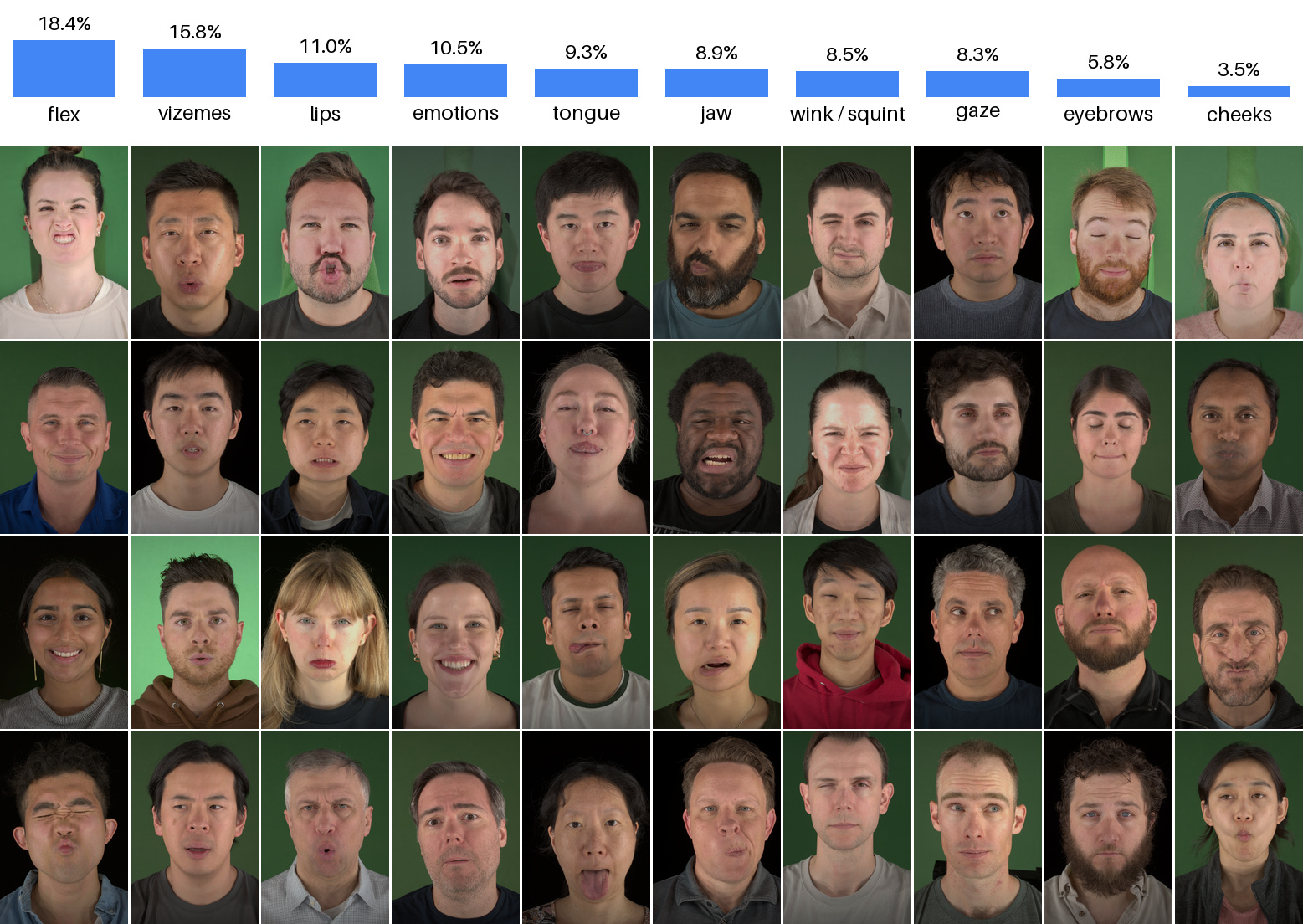}
    \caption{Dataset expression categories and distribution. A breakdown of the sample distribution across ten facial expression categories, alongside visual examples demonstrating the specific movements and demographic diversity captured for each group.}
    \label{fig:dataset_expression_samples}
\end{figure}

The dataset contains over $\sim5\,000$ individuals covering a diverse demographic background, see Figure~\ref{fig:hb_stats}. Each individual performs a set of static facial expressions producing $\sim150\,000$ samples in total. The capture protocol defines anatomically global and local facial activations which can be grouped into the following categories: flexing (e.g. stretching and compressing the face, smiling), standard visemes (10 categories), lips motion (e.g. rolling, pucker, funneler), global emotions (e.g. sadness, fear), tongue motions (e.g. rolling, sideway motions), jaw motions (sideway motion), winking and squinting (with single and both eyes), gaze (changing vertical and horizontal gaze direction), eyebrows motion (raising and lowering) and cheeks deformation (sucking and blowing), see Figure~\ref{fig:dataset_expression_samples}. To process this dataset of raw multi-view stereo reconstruction at scale, we developed a custom, highly parallelized data processing pipeline, optimized to process $\sim10\,000$ samples per day. 

\subsection{Head Registration}
\label{sec:face_registration}

Following FLAME \cite{li2017learning}, GNM employs an iterative coregistration cycle \cite{hirshberg2012coregistration} alternating between face registration and statistical model building. Initialized with a custom parametric model derived from curated 3D head meshes, it registers a large multi-view dataset to produce new registrations that continually refine the model.

Our multi-stage pipeline first fits the GNM to the scan, then non-rigidly deforms the mesh surface via per-vertex offsets. This relies on unshaded inverse rendering of the template mesh and its RGB texture, implemented in Mitsuba \cite{Mitsuba3} using edge sampling \cite{Li2018} for visibility gradients. The optimization jointly solves for GNM parameters, 3D offsets, and the RGB texture by minimizing image-based losses and geometric priors. To mitigate tangential sliding across expressions, we initially register a subject-specific neutral scan. Non-neutral registrations then initialize with this neutral texture and minimize a UV-space SSIM loss between the two.

\textbf{Image-based losses}. Alongside RGB supervision, we extract auxiliary signals from the captured images, dense face landmarks \cite{wood20223d}, a normal buffer from a custom multi-view stereo reconstruction \cite{qiu2025chosen}, and per-pixel semantic segmentation. Our renderer outputs corresponding normal and semantic Arbitrary Output Variables (AOVs). We apply an L1 loss between these renderings (RGB, normal, semantic) and their ground truths, alongside an SSIM loss on the RGB output to promote camera-space alignment. Normal supervision enforces accurate surface orientation. Semantic supervision aligns visible facial structures such as the eyes, ears, and lips. For frequently occluded regions such as the teeth and tongue, we primarily rely on dense landmarks.

\textbf{Geometry regularization}. For geometric stability, we apply a gradient descent preconditioner \cite{Nicolet2021Large}, L2 regularization on per-vertex offsets, and minimize their graph Laplacian's L2 norm \cite{taubin1995}. We also penalize edge deviations between the fitted GNM and displaced vertices \cite{li2017learning}. Finally, a custom differentiable loss prevents self-intersections in high-curvature regions such as the ears, tongue, and lips.

\begin{figure}
    \centering
    \includegraphics[width=1.0\linewidth]{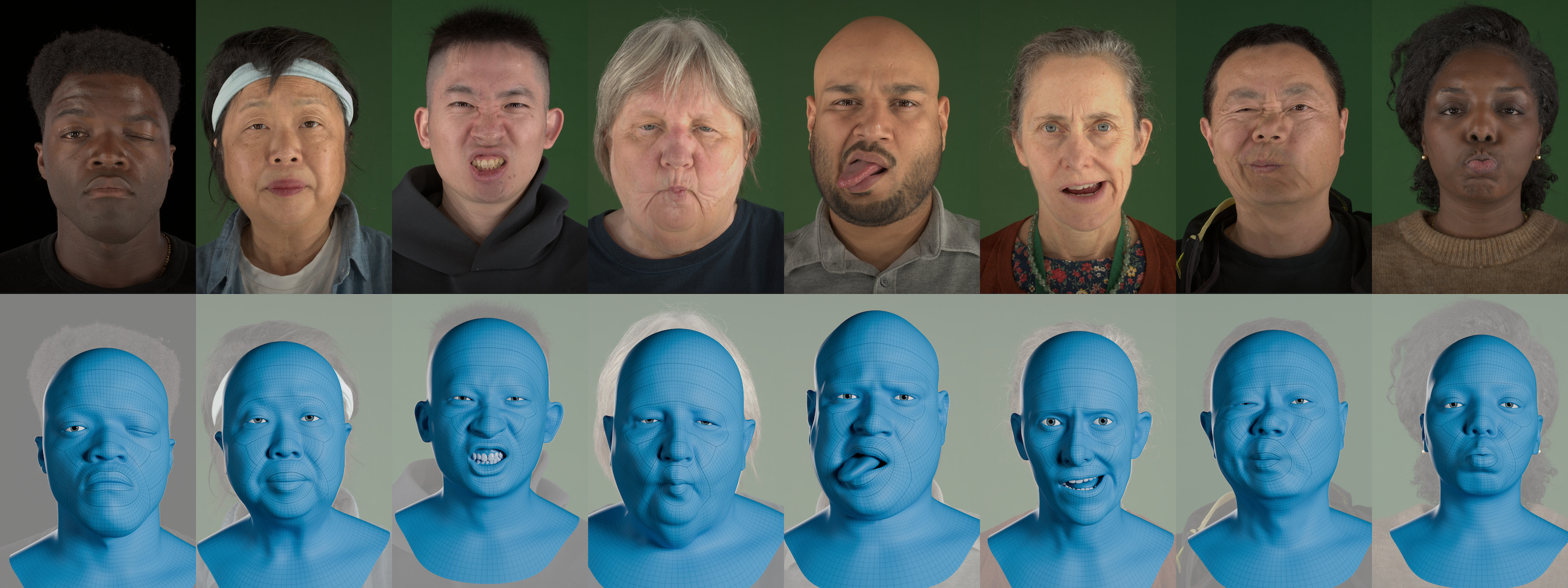}
    \caption{A set of example registrations showcasing the ability of our registration pipeline to reconstruct highly deformed expressions and intraoral movements. Top row: single raw images from 3D scan data. Bottom row: Corresponding registration meshes, illustrating precise surface tracking and mathematically decoupled alignment of the external facial surface alongside internal structures.}
    \label{fig:hb_registrations}
\end{figure}

While the capture system reliably reconstructs the frontal face, hair occlusion prevents direct empirical measurement of the cranium. To reconstruct an anatomically plausible head without approximations such as Laplacian smoothing, we employ a cross-domain latent-space regression strategy \cite{ploumpis2019combining}. We build an auxiliary model from 200 accurate, artist-sculpted head meshes. Projecting our registered faces onto this model yields a physiologically plausible cranial shape while retaining detailed facial features. A final non-rigid deformation maps the mesh back to the original registration, preserving the newly acquired cranium.

We build GNM from these final registration meshes (Section \ref{sec:gnm_formulation}); some example reconstructions can be seen in Figure~\ref{fig:hb_registrations}. The pipeline reconstructs the external skin alongside internal structures (eyeballs, teeth, and tongue) even under severe occlusion. This iterative coregistration relies on the GNM as a geometric prior to accurately deform internal components from sparse signals such as visible teeth landmarks. To bootstrap the model, we initially integrated artist-sculpted internal parts into the base template and its corresponding spaces (Section \ref{sec:initial_placement_of_teeth_and_tongue}).

\subsection{Model Formulation} \label{sec:gnm_formulation}

GNM is formulated as a function $\gnmfunc(\gnmparams; \gnmdata): \real^{|\gnmparams|} \rightarrow \real^{\ngnmverts \times \gnmvertdim}$, which produces a human head mesh of $\ngnmverts$ $\gnmvertdim$D vertices given a set of model parameters $\gnmparams$ and fixed model data $\gnmdata$. The model parameters ${\gnmparams = \left(\paramsid, \paramsexpr, \paramsrot, \paramstransl\right)}$ consist of identity $\paramsid \in \real^{\nparamsid}$ parameters, expression $\paramsexpr \in \real^{\nparamsexpr}$ parameters, angle-axis rotations of $\ngnmjoints=4$ joints $\paramsrot \in \real^{\nparamsrot}$ and global translation $\paramstransl \in \real^{\nparamstransl}$. The model data $\gnmdata = \left(\gnmverttempl, \gnmjointstempl, \idbasis, \exprbasis, \jointsbasis, \skinningweights, \jointparents \right)$ consist of a template head mesh $\gnmverttempl \in \real^{\ngnmverts \times \gnmvertdim}$, template joint locations $\gnmjointstempl \in \real^{\ngnmjoints \times \gnmvertdim}$, identity basis $\idbasis \in \real^{\nparamsid \times \gnmvertdim \times \ngnmverts}$, expression basis $\exprbasis \in \real^{\nparamsexpr \times \gnmvertdim \times \ngnmverts}$, joint location identity basis $\jointsbasis \in \real^{\nparamsid \times \gnmvertdim \times \ngnmjoints}$, LBS weights $\skinningweights \in \real^{\ngnmjoints \times \ngnmverts}$ and the kinematic chain definition $\jointparents \in \integers^{\ngnmjoints}$.

The model function is defined as $\gnmfunc(\gnmparams; \gnmdata) = \lbsfunc\left(\vertsbind, \skinningtf; \blendweights \right)$, where $\lbsfunc$ is a standard LBS function, which rotates and blends the bind pose vertices $\vertsbind \in \real^{\ngnmverts \times \gnmvertdim}$ by the skinnning transformations $\skinningtf~\in~\real^{\ngnmjoints\times 4 \times 4}$ and blendweights $\blendweights$. The bind pose vertices are computed as $\vertsbind = \idexprfunc(\paramsid, \paramsexpr; \gnmverttempl, \idbasis, \exprbasis)$, where the function $\idexprfunc: \real^{\nparamsid \times \nparamsexpr} \rightarrow \real^{\ngnmverts \times \gnmvertdim}$ applies per-vertex identity and expression offsets to the template $\gnmverttempl$. Formally,
\begin{align} \label{eq:gnm_feedforward}
\idexprfunc(\paramsid, \paramsexpr; \gnmverttempl, \idbasis, \exprbasis) = \gnmverttempl + \sum_{i}^{\nparamsid}{\paramsidi{i} \idbasis_{i}} + \sum_{i}^{\nparamsexpr}{\paramsexpri{i}\exprbasis_{i}}.
\end{align}
The skinning transforms are computed as $\skinningtf = \skinningtffunc(\jointsbind, \paramsrot, \paramstransl; \jointparents)$, where the function $\skinningtffunc: \real^{|\jointsbind| \times |\paramsrot| \times \nparamstransl} \rightarrow \real^{\ngnmjoints \times 4 \times 4}$, takes the bind pose joint locations $\jointsbind \in \real^{\ngnmjoints \times \gnmvertdim}$ and propagates the global translation $\paramstransl$ and per-joint rotations $\paramsrot$ through the kinematic chain $\jointparents$ to obtain global per-joint affine transforms.

The bind pose joint locations are computed as  $\jointsbind = \jointsbindfunc(\paramsid; \gnmjointstempl, \jointsbasis)$, where the function $\jointsbindfunc: \real^{\nparamsid} \rightarrow \real^{\ngnmjoints \times \gnmvertdim}$ is defined as
\begin{align}
\jointsbindfunc(\paramsid; \gnmjointstempl, \jointsbasis) = \gnmjointstempl + \sum_{i}^{\nparamsid}{\paramsidi{i}\jointsbasis_{i}}.    
\end{align}
Finally, the $i$-th output mesh vertex $\vertex^{(i)} \in \real^{3}$ is computed as
\begin{align}
    \lbsfunc^{(i)}\left(\vertsbind, \skinningtf; \blendweights \right) = \sum_{k=1}^{\ngnmjoints}{\homogfuncinv\left( \blendweights_{k,i} \skinningtf_{k} \homogfunc\left(\transpose{\vertsbind_{i}} \right) \right)},
\end{align}
where $\homogfunc: \real^{3} \rightarrow \real^{4}$, $\homogfunc(\vertex) = \bigl( \begin{smallmatrix} \vertex \\ 1 \end{smallmatrix} \bigr)$ converts a 3D vector into homogeneous coordinates, and $\homogfuncinv$ does the opposite.

\begin{figure}
    \centering
    \includegraphics[width=0.85\textwidth]{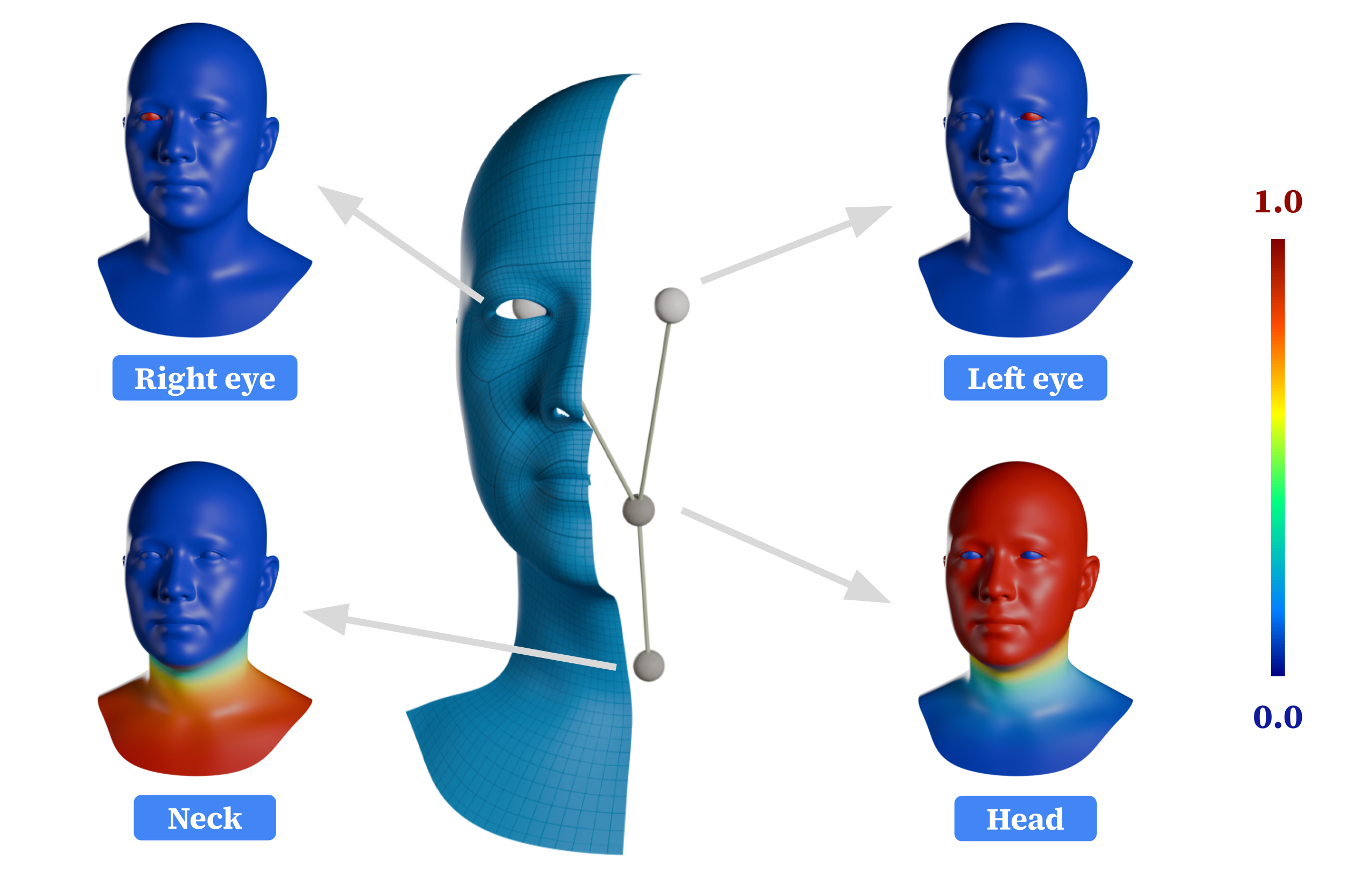}
    \caption{Template mesh and skinning blendweights. The central model illustrates the neutral head template. The surrounding heatmaps display the manually designed blendweights for the four articulated joints: the right eye, left eye, neck, and head. The colour scale indicates the degree of joint influence over the mesh, ranging from 0.0 (blue, no influence) to 1.0 (red, full influence).}
    \label{fig:kinematic_tree}
\end{figure}

The template head mesh $\gnmverttempl$, identity basis $\idbasis$, expression basis $\exprbasis$ and joint location basis $\jointsbasis$ are learned from the data, whereas the skinning blendweights $\blendweights$ and the kinematic chain $\jointparents$ are designed manually by an artist, see Figure~\ref{fig:kinematic_tree}. The template joint locations $\gnmjointstempl$ rely both on an artist defined joint regressor and the data driven identity basis $\idbasis$.

\subsection{Composite Linear Bases} \label{sec:composite_linear_bases}
To achieve higher fidelity of the shape space covered by the model, and to allow for the use of highly-specialized datasets, we split the identity and expression bases $\idbasis$ and $\exprbasis$ into portions which correspond to anatomical regions of the human head.

Specifically,
\begin{align}
    \idbasis = \idbasishead~\big\|~\idbasiseyeballs~\big\|~\idbasisteeth,
\end{align}

where $\idbasishead, \idbasiseyeballs, \idbasisteeth \in \real^{\nparamsidregion{r} \times \ngnmverts \times \gnmvertdim}$ are the head, teeth and eyeballs portions of $\idbasis$ stacked along the first tensor dimension, and where $\nparamsidregion{r}$ represents the number of components in portion $r \in \{\text{\headregion}, \text{\eyeballregion}, \text{\teethregion}\}$. Note, that $\paramsid = \transpose{\left[ \paramsidheadt~\paramsideyeballst~\paramsidteetht \right]}$.

Similarly, 
\begin{align}
    \exprbasis = \exprbasislefteye~\big\|~\exprbasisrighteye~\big\|~\exprbasislowerface~\big\|~\exprbasistongue~\big\|~\exprbasispupil,
\end{align}

where $\exprbasislefteye, \exprbasisrighteye, \exprbasislowerface, \exprbasistongue, \exprbasispupil \in \real^{\nparamsexprregion{r} \times \ngnmverts \times \gnmvertdim}$ are the left and right periocular region, lower face, tongue and eyeball pupil portions of $\exprbasis$, and where $\nparamsexprregion{r}$ represents the number of components in portion $r \in \{\text{\lefteyeregion}, \text{\righteyeregion}, \text{\lowerfaceregion}, \text{\tongueregion}, \text{\pupilregion}\}$. Note, that $\paramsexpr = \transpose{\left[ \paramsexprlefteyet~\paramsexprrighteyet~\paramsexprlowerfacet~\paramsexprtonguet~\paramsexprpupilt \right]}$.

The definition of each of the portions of $\idbasis$ and $\exprbasis$ is detailed in the following sections.

\subsection{Head Identity} \label{sec:identity}
Here we define the head identity basis $\idbasishead$, the template mesh $\gnmverttempl$, the template joint locations $\gnmjointstempl$ and the joint location identity basis $\jointsbasis$. The head identity basis $\idbasishead$ is computed as a standard PCA over the set of neutral face meshes. As $\idbasishead$ should only capture the deformation due to the change of human identity and not due to a rigid head movement in space, we first rigidly align all the neutral face meshes to a common template using Procrustes alignment \cite{luo2002iterative}, obtaining the dataset $\fulldatasetneutral \in \real^{\nsamplesdatasetneutral \times \gnmvertdim\ngnmverts}$ of $\nsamplesdatasetneutral$ samples representing flattened $\gnmvertdim$D mesh vertices.

While $\fulldatasetneutral$ contains all $\ngnmverts$ vertices of the GNM topology, we zero-out the vertices corresponding to the eyeballs, and place them in the final model manually to achieve a higher fidelity of the eyeball-eyelid contact. We denote this modified dataset $\datasetneutral$. First, we explain how we compute the identity basis $\headskinidbasis \in \real^{\nparamsid \times \ngnmverts \times \gnmvertdim}$ which is equivalent to $\idbasishead$ except for the  zeroed-out eyeball vertices. Next, we explain how we inject the eyeball deformation caused by the identity change into $\headskinidbasis$ to obtain the final $\idbasishead$.

We compute $\idbasisvectors, \idbasiseigvals = \eig{\covmatid}$, where $\idbasisvectors \in \real^{\nidbasisvectors \times \gnmvertdim\ngnmverts}$ and $\idbasiseigvals \in \real^{\nidbasisvectors}$ are the eigenvectors and their associated eigenvalues and $\covmatid \in \real^{\gnmvertdim\ngnmverts \times \gnmvertdim\ngnmverts}$ is a covariance matrix of the centered dataset $\centereddatasetneutral = \datasetneutral - \ones\datasetneutralmean$, where $\datasetneutralmean = \frac{1}{\nsamplesdatasetneutral}\sum_{i=1}^{\nsamplesdatasetneutral}{\datasetneutral_{i}}$ is the dataset mean and $\ones$ is a vector of ones. Let $\flattenfunc: \real^{a \times b} \rightarrow \real^{ab}$ be a function that flattens a matrix to a vector, while $\flattenfuncinv$ does the opposite. The $i$-th component of $\headskinidbasis$ is computed as $\headskinidbasisi{i} = \flattenfuncinv \left(\idbasisvectors_{i}\idbasiseigvalsi{i}\right)$, that is, each basis vector is scaled proportionally to the dataset variance it explains, which unifies the effective range of the identity coefficients $\paramsid$. The basis vectors are thus orthogonal but not unit-length. Note, that we only keep the first $\nparamsidhead \ll \nidbasisvectors$ basis vectors which jointly explain $\sim99\%$ of the dataset variance.

We express the GNM head template mesh as $\gnmverttempl = \flattenfuncinv(\datasetneutralmean)$, which we further modify by replacing the teeth by the mean of the teeth dataset $\datasetteeth$ of Section \ref{sec:tongue_modeling_and_articulation}, and by optimizing eyeball location to best fit the eyelids.

To backfill the eyeballs in the identity basis, we optimize a uniform rigid translation for the eyeball vertices in each basis component to maintain a plausible eyeball-eyelid contact under identity deformations. For each basis component $\headskinidbasisi{i}$, we consider the positive and negative bound deformations of the eyelid vertices at a fixed magnitude and find an optimal translational offset such that the displaced eyeballs best fit the corresponding eyelid vertices of the positive and negative deformations. Thus we ensure that as the face identity changes, the eyeballs translate to stay aligned with the deforming eyelids. This modification results in the final $\idbasishead$.


The template joints $\gnmjointstempl$ were placed in the template mesh $\gnmverttempl$ manually by an artist, whereas the joint location identity basis $\jointsbasis$, which ensures that the skeleton scales and moves appropriately with each subject's unique head shape controlled by the identity parameters $\paramsid$, is computed automatically using a linear joint regressor $\jointregressor \in \real^{\ngnmjoints \times \ngnmverts}$. Specifically, the $i$-th joint component $\jointsbasis_{i} = \jointregressor \idbasis_{i}$. The joint regressor itself is computed as an optimization problem $\jointregressor = \argmin_{\jointregressor}{\frob{\jointregressor\gnmverttempl - \gnmjointstempl} + \mathcal{L_{\text{reg}}}}$, where $\mathcal{L_{\text{reg}}}$ is a regularizer which encourages sparsity and left-right mesh symmetry.

\subsection{Head Expression} \label{sec:expression}

Here we define the left and right periocular region and the lower face portions $\exprbasislefteye$, $\exprbasisrighteye$ and $\exprbasislowerface$ of $\exprbasis$. Each region is computed as an uncentered PCA over the set of samples representing vertex displacements of expressive faces w.r.t. the neutral one for each human subject. Similarly to $\idbasis$, $\exprbasis$ should only capture shape variation due to a facial expression change, but not any rigid face motion. Therefore, we first align the registered meshes using a face mesh stabilization technique described in more detail at the end of this section.

Let $\vertssubjneut{s}, \vertssubjexpr{s}{i} \in \real^{\ngnmverts \times \gnmvertdim}$ be the neutral and the $i$-th expressive face mesh of subject $s$, where $\vertssubjexpr{s}{i}$ is stabilized to $\vertssubjneut{s}$, and let $\exprdelta{s}{i} \in \real^{\gnmvertdim\ngnmverts}, \exprdelta{s}{i} = \flattenfunc\left( \vertssubjexpr{s}{i} - \vertssubjneut{s} \right)$ represent a corresponding flattened array of per-vertex deltas. Let $\nsamplesdatasetexpressive$ be the total number of dataset samples, $\nsubjects$ be the total number of unique subjects and $\nsubjectsamples{s}$ be the number of expressive faces available for a subject $s$. The expression dataset $\datasetexpressive \in \real^{\nsamplesdatasetexpressive \times \gnmvertdim\ngnmverts}$, is then defined as $\datasetexpressive = \transpose{\left[ \exprdelta{1}{1} \dots \exprdelta{1}{\nsubjectsamples{1}} \exprdelta{2}{1} \dots \exprdelta{2}{\nsubjectsamples{2}} \dots \exprdelta{\nsubjects}{1} \dots \exprdelta{\nsubjects}{\nsubjectsamples{\nsubjects}} \right]}$. Any deformation of the tongue or eyeballs is zeroed-out, as we compose their expression basis individually, see Sections \ref{sec:tongue_modeling_and_articulation} and \ref{sec:eyeball_model_formulation}. However, note that the motion of the jaw together with the lower teeth is modeled  fully within $\exprbasis$.

\begin{figure}
    \centering
    \includegraphics[width=0.85\textwidth]{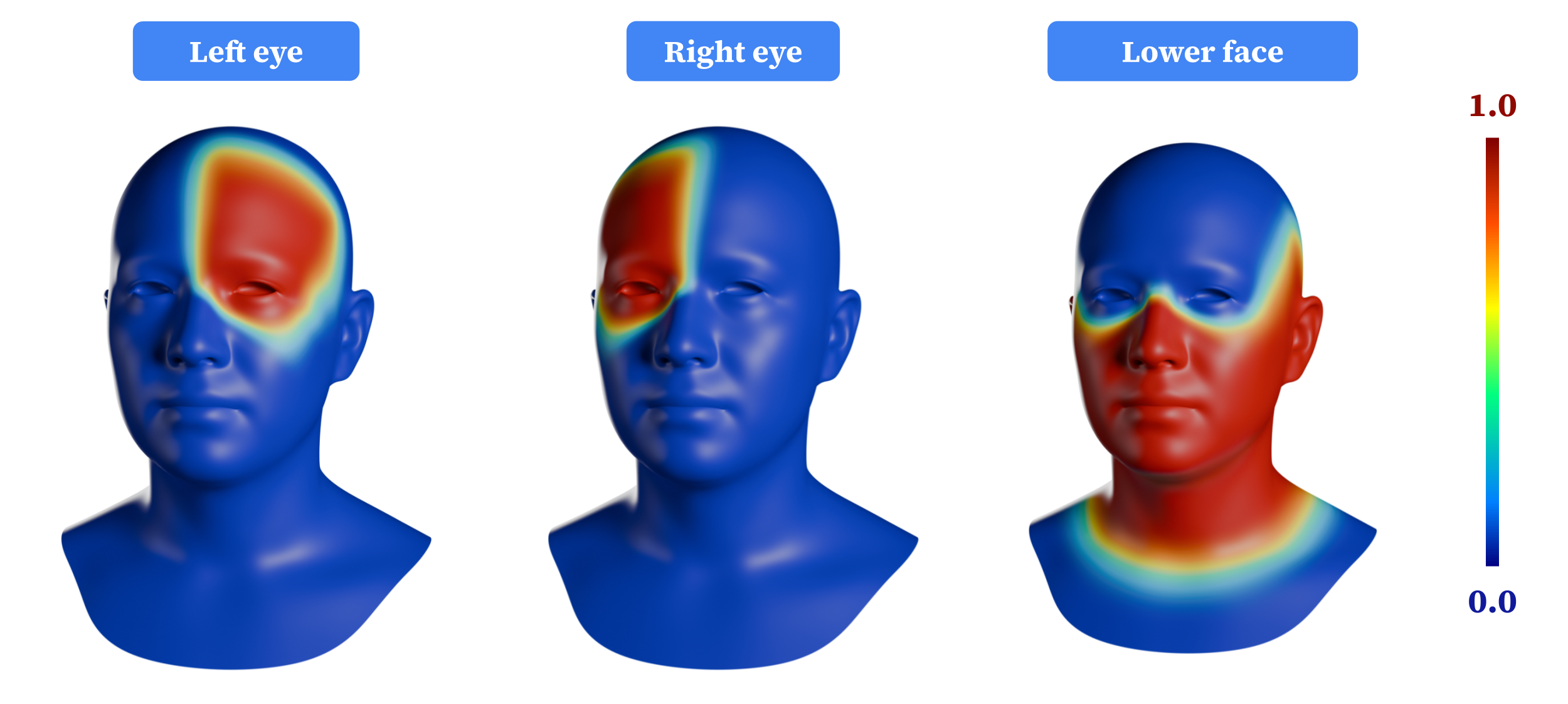}
    \caption{Regional expression basis masks. Visualisation of the continuous vertex masks ($S_r$) used to spatially partition the head mesh into three distinct areas: the left periocular region, the right periocular region, and the lower face. Warmer colours indicate higher mask weights ($S_r \to 1$). This regional formulation isolates the computation of the expression basis, yielding two critical benefits: it provides localised, intuitive control (preventing semantic leakage, such as a jaw movement triggering an eye wink), and it ensures the left and right periocular expressions can be perfectly mirrored.}
    \label{fig:expression_heatmap}
\end{figure}

Dividing the head mesh into the left and right periocular regions and the lower face region (see Figure~\ref{fig:expression_heatmap}) for the sake of computing the expression basis comes with two benefits: (i) the model allows for localized and more intuitive control of a facial expression (e.g. an opening mouth will not trigger an eye wink), and (ii) the left and right periocular region expressions are perfectly mirrored. Let $\vertexmaskregion{r} \in \real^{\gnmvertdim\ngnmverts}$ be a vertex mask of a region $r$ with values in $[0, 1]$. We then define a per-region expression dataset $\datasetexpressiveregion{r}$, composed of samples $\datasetexpressiveregion{r}_{i} = \datasetexpressive_{i}\vertexmaskregion{r}$.

We compute $\exprbasisvectorsregion{r}, \exprbasiseigvalsregion{r} = \eig{\covmatexprregion{r}}$, where $\exprbasisvectorsregion{r} \in \real^{\nexprbasisvectorsregion{r} \times \gnmvertdim\ngnmverts}$ and $\exprbasiseigvalsregion{r} \in \real^{\nexprbasisvectorsregion{r}}$ are the eigenvectors and their associated eigenvalues, $\nexprbasisvectorsregion{r}$ is the number of eigenvectors computed for region $r$, and $\covmatexprregion{r} \in \real^{\gnmvertdim\ngnmverts \times \gnmvertdim\ngnmverts}$ is a covariance matrix of \textit{uncentered} dataset $\datasetexpressiveregion{r}$. We avoid centering $\datasetexpressiveregion{r}$ since modifying the template mesh $\gnmverttempl$ to absorb the mean of $\datasetexpressiveregion{r}$ would result in a face with slightly closed eyes and a slightly opened mouth. Instead, to make controlling GNM intuitive, we design the model so that zero identity and expression parameters $\paramsid, \paramsexpr$ produce a perfectly neutral face.

The $i$-th component is computed as $\exprbasisregioni{i}{r} = \flattenfuncinv\left( \exprbasisvectorsregioni{i}{r}\exprbasiseigvalsregioni{i}{r} \right)$ so as to scale the components proportionally to the explained variance, and we keep only the first $\nparamsexprregion{r} \ll \nexprbasisvectorsregion{r}$ which jointly explain $\sim99\%$ of the dataset variance. The expression components $\exprbasis^{(r)}$ are orthogonal within each region $r$, but not among the regions, as the vertex masks of the neighboring regions have small overlaps, where the contribution of the per-region components are linearly blended. Note, that we only compute $\exprbasisregion{left\_eye}$, and then mirror the individual displacement vectors along a vertical plan splitting the head into symmetric left and right halves, to obtain $\exprbasisregion{right\_eye}$.

As mentioned above, to remove any spurious misalignment within the source dataset, we first stabilize the face meshes. Stabilization finds a rigid 6-DOF transform between a source and a target mesh representing a skin surface of the same human subject, so that the underlying (and unknown) skull aligns perfectly in space. It is a notoriously difficult vision problem \cite{beeler2014rigid,wu18incremental}, therefore, to achieve a precise alignment, we take a two stage approach.

First, we apply fully automatic modified confidence map stabilization \cite{bednarik2024learning}. Then, any remaining misalignments are removed by a semi-automatic PCA-based stabilization, which operates as follows. We define five granular facial regions (left and right periocular region, nose, mouth, neck with back of the head), and perform a PCA decomposition for each, conceptually taking the same steps as when computing $\exprbasisregion{r}$. The assumption is that any strong spurious rigid 6-DOF deformation would get naturally contained in a few PCA components. We visualize the contribution of the individual per-region components, a human operator visually identifies and discards the ones representing an unwanted deformation, and we rebuild the data from the remaining ones, thus creating a clean set of expressive meshes to build $\datasetexpressive$ from.

\subsection{Internal Anatomy} \label{sec:internal_anatomy}
To make GNM truly comprehensive, internal anatomy must be included too. Modeling the teeth, tongue and eyeballs poses a significant challenge due to the data scarcity, complex teeth geometry, and a high degree of freedom in the motion of a tongue. We address these limitations by leveraging a hybrid approach that combines artist-guided synthetic models and diverse real-world 3D scans.

\subsubsection{Teeth modeling}
GNM integrates a parametric dental subsystem that replaces static, generic mouth templates a shape controlled by the teeth identity basis $\idbasisteeth$ and the corresponding parameters $\paramsidteeth$. The model is designed to represent diverse dental arches, accurately capturing natural variations in teeth shape, size, and individual alignment while remaining compatible with the overall head shape.

Starting from a generic template teeth model, an artist rigged each tooth and procedurally generated $\nsamplesdatasetteeth = 5\,000$ synthetic dental shapes, including both the upper and lower teeth along with their associated gums. We denote this dataset $\datasetteeth \in \real^{\nsamplesdatasetteeth \times \gnmvertdim\ngnmverts}$. Note that all the vertices except for teeth are set to $0$. As before,  $\idbasisteeth$ is computed as a standard PCA on the dataset $\datasetteeth$, where the eigenvectors are scaled by their respective eigenvalues represent the final $\idbasisteeth$. We again only keep the first $\nparamsidteeth$ components which explain $\sim99\%$ of the dataset variance.

\subsubsection{Tongue Modeling and Articulation} \label{sec:tongue_modeling_and_articulation}
We develop a tongue model to address the lack of expressiveness in the inner-cavity in traditional 3DMMs. The tongue model is represented as the portion $\exprbasistongue$ of the expression basis $\exprbasis$, and it is, too, computed as a PCA decomposition of a dataset of vertex displacements w.r.t. the neutral template.

 The tongue represents a highly deformable surface which is notoriously difficult to register from raw scans \cite{ploumpis20223d}. To capture the broad tongue shape space, we generate a specialized dataset by fitting a custom artist-made tongue rig to two data sources. First, we take a subset of expressive faces performing specific tongue movements (sideway motion, rolling etc.) from our registrations described in Section \ref{sec:face_registration}, we only keep the outer skin part of the face and use our tongue rig to generate varied synthetic tongue poses by respecting the overall face and lip geometry to avoid any penetrations. Second, we fit the tongue rig to the dataset of sparse 3D keypoints from \cite{medina2022speech} while applying rigorous constraints to the tongue’s global morphology and curvature to ensure anatomical plausibility.

As in Section \ref{sec:expression}, we compute per-vertex displacements to the template, and denote the resulting dataset as $\datasettongue \in \real^{\nsamplesdatasettongue \times \gnmvertdim\ngnmverts}$, where $\nsamplesdatasettongue \sim 2.5\textrm{K}$ is the dataset size. Note, that all the vertices except for the tongue are set to $0$. As before, $\exprbasistongue$ is computed as a standard (centered) PCA on the dataset $\datasettongue$, where the eigenvectors $\exprbasisvectorsregion{tongue}$ scaled by their respective eigenvalues $\exprbasiseigvalsregion{tongue}$ represent the components of $\exprbasistongue$. As before, we only keep the first $\nparamsexprregion{tongue} \ll \nexprbasisvectorsregion{\text{tongue}}$ components explaining $\sim 99\%$ of the dataset variance.

Note, that the mean $\datasettonguemean$ of the dataset $\datasettongue$ must be included as a summand in Eq.~\ref{eq:gnm_feedforward} producing the posed head vertices. An intuitive solution would be to absorb $\datasettonguemean$ in $\gnmverttempl$. However, as $\exprbasistongue$ is computed independently of the rest of the face, $\datasettonguemean$ represents a tongue protruding the surface of a neutral face mesh. Therefore, we instead absorb $\datasettonguemean$ as the first component of $\exprbasistongue$. Setting $\paramsexpr$ to the default value of $\zeros$ thus leads to a retracted tongue tucked away inside the mouth cavity.

\subsubsection{Eyeball Model Formulation} \label{sec:eyeball_model_formulation}

\begin{figure}
\centering

\newcommand{\leftwidth}{0.40\textwidth}
\newcommand{\gapwidth}{0.05\textwidth}
\newcommand{\rightwidth}{\dimexpr\textwidth - \leftwidth - \gapwidth\relax} 

\begin{tabular}{@{} c @{\hspace{\gapwidth}} c @{}}
  
  \adjustbox{valign=c}{\includegraphics[width=\leftwidth]{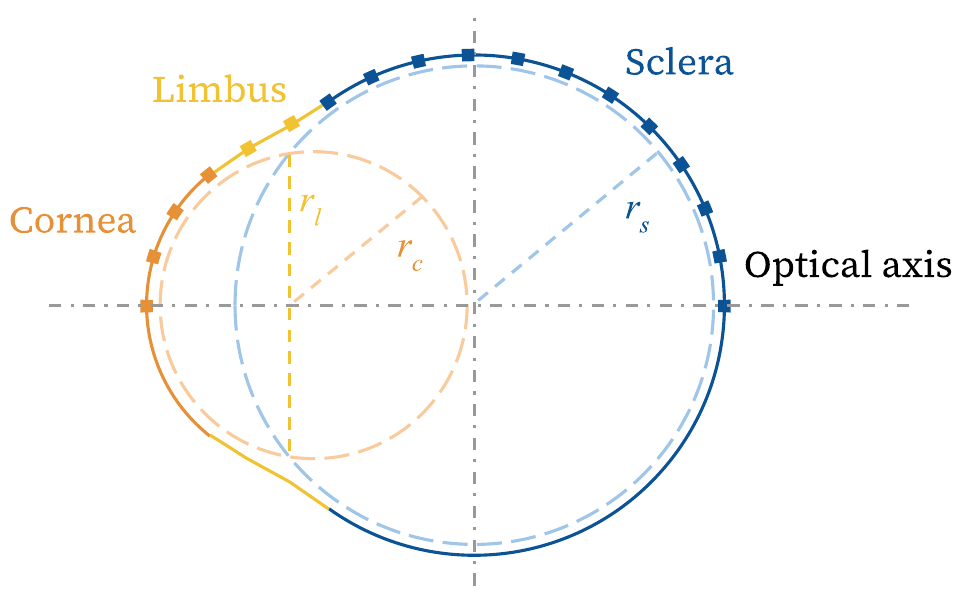}} &
  \adjustbox{valign=c}{\includegraphics[width=\rightwidth]{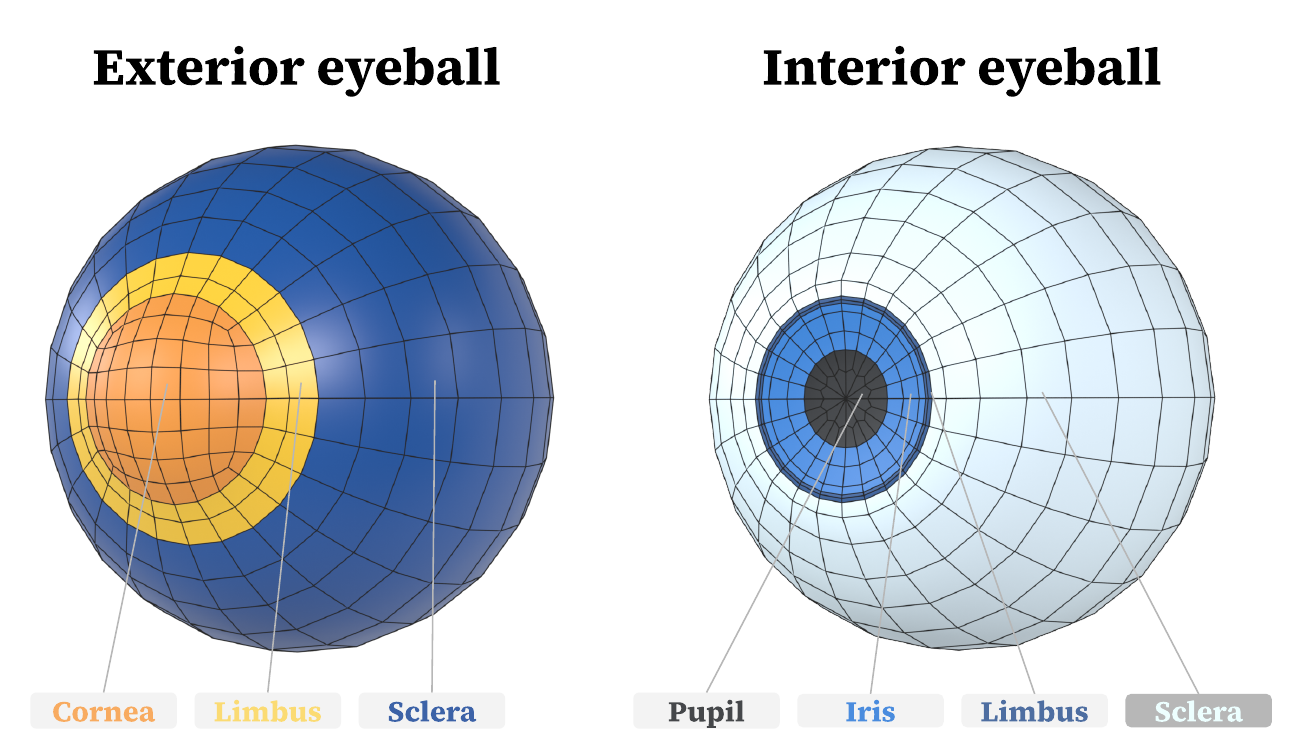}} \\
  \\[-1ex]  
  
  \begin{subfigure}[t]{\leftwidth}
    \captionsetup{justification=centering}
    \caption{Cross section view (exaggerated proportions)}
    \label{fig:eyeball_cross_section}
  \end{subfigure} &
  \begin{subfigure}[t]{\rightwidth}
    \captionsetup{justification=centering}
    \caption{Geometry schematic}
    \label{fig:eyeball_geometry}
  \end{subfigure}

\end{tabular}
\caption{Parametric representation of the GNM ocular geometry. (a) A 2D cross-sectional view illustrating the underlying two-sphere model. The optical axis connects the centres of the scleral and corneal spheres. To maintain anatomical compatibility with the eyelids, the scleral radius ($r_s$) is held constant, whilst the corneal ($r_c$) and limbal ($r_l$) radii are sampled from physiological distributions. (b) 3D schematics of the exterior and interior eyeball meshes. The linear identity basis ($\idbasiseyeballs$), derived from the 2D cross-section polylines, is interpolated onto these 3D surfaces as a surface of revolution. The interior mesh highlights the pupil, which is independently driven by a dedicated expression basis ($\exprbasispupil$) to enable dynamic, lighting-dependent dilation.}
\label{fig:eyeball}
\end{figure}

We define the eyeball as a linear parametric model designed to bridge the gap between simplified spherical representations and the complex, person-specific ocular geometry required for high-fidelity gaze tracking. The model represents the $\idbasiseyeballs$ portion of the identity basis $\idbasis$ and the $\exprbasispupil$ portion of the expression basis $\exprbasis$.

Accurate cornea modeling is particularly critical for synthetic data generation, as the subtle curvature of the cornea dictates the formation of LED glints, which are primary features in gaze-target prediction \cite{guestrin2006general}. By integrating corneal and limbal parameters directly into the identity space $\idbasis$, we ensure that every generated identity possesses biologically plausible ocular features compatible with the overall facial topology.

The eyeball geometry is defined as a two-sphere model where a large scleral sphere of radius $r_s$ forms the bulk of the eye, and a smaller corneal sphere of radius $r_c$ defines the anterior segment, see Figure \ref{fig:eyeball_cross_section}. The optical axis is established by the line connecting these two sphere centers. To capture human anatomical diversity while maintaining eyelid compatibility with the GNM template, the scleral radius is held constant ($r_s \approx 14.6\text{ mm}$). Conversely, the remaining parameters are drawn from established physiological Gaussian distributions--specifically for the limbus radius $r_l$ ($\mu = 6.0\text{ mm}, \sigma = 0.44\text{ mm}$) and cornea radius $r_c$ ($\mu = 8.5\text{ mm}, \sigma = 0.73\text{ mm}$)--to yield a pseudo-ground truth dataset.

To transform these non-linear geometric properties into a linear basis $\idbasiseyeballs$, we first sample a large set ($N=\textrm{10\,000}$) of 2D eyeball cross-sections as polylines. These polylines are generated by varying the nonlinear parameters ($r_l$, $r_c$) according to their physiological statistics. Importantly, a smoothing algorithm is applied to the transition zone between the sclera and cornea in each polyline to achieve a more organic and continuous surface, avoiding sharp geometric discontinuities. PCA is then applied to the vertices of these smoothed polylines to extract the leading modes of shape variation. The resulting 2D linear identity basis is subsequently transferred to the 3D template eyeball interior and exterior mesh vertices via interpolation in polar coordinates, exploiting the eyeball's nature as a surface of revolution. This process allows the eyeball’s shape to be controlled by a small set of identity coefficients, seamlessly integrated into the primary GNM identity basis.

The eyeball expression basis $\exprbasispupil$ is focused solely on pupil dilation. We employ a hand-crafted geometric basis parameterized such that a single coefficient, scaled to operate within a range of $[-3, 3]$, controls pupil dilation. At a coefficient value of -3, the pupil radius contracts to a point, at 0 it is half the radius of the iris, and at +3, it expands to match the full iris size. This range comfortably encompasses biologically sensible radii (e.g., 1 to 4 mm) and provides essential variability for lighting-dependent gaze models or expressive rendering. Both eyeballs of a GNM instance share the same shape identity $\idbasiseyeballs$ and expression $\exprbasispupil$ parameters, ensuring bilateral symmetry. Anatomical coherence with the eyelids is primarily established during the GNM model construction phase: the main head identity basis components are analyzed for their impact on the eye socket shape, and corresponding rigid transformations are computed and integrated into both the vertex and joint identity bases for all eyeball vertices and their rotation pivots, ensuring the eyeballs translate appropriately with facial identity changes.


\subsection{Initial Placement of Teeth and Tongue} \label{sec:initial_placement_of_teeth_and_tongue}

As discussed in Section \ref{sec:face_registration}, the initial version of the GNM model had artist-designed teeth, tongue and eyeballs manually injected in the identity and expression basis, after it is computed from the registered data. Here we describe this process.

To backfill the teeth and tongue in the identity basis $\idbasis$, we assume that any face deformation caused by an identity change can lead to a rigid 3-DOF translation of the teeth and tongue. As an approximation, we observe the mean displacement of the inner lower and upper lip vertices induced by each component $\idbasis_{i}$ and store this value into the (thus far zeroed-out) portion of $\idbasis_{i}$ corresponding to the teeth and tongue vertices.

To backfill the lower teeth and tongue in the expression basis $\exprbasis$, we first note that the any facial expression change which moves the jaw, rigidly transforms the lower teeth too. As an approximation, we define a vertex group covering the lower lip and the chin, and we find a rigid 6-DOF transform of these vertices induced by each component $\exprbasis_{i}$ using a Procrustes alignment. The recovered transform is applied to all the lower teeth and tongue vertices in $\exprbasis_{i}$.

\subsection{Model Implementation Details} \label{sec:implementation_details}
As explained in section \ref{sec:composite_linear_bases}, the identity and expression bases $\idbasis$ and $\exprbasis$ are composed of disjoint anatomical regions, each represented as a PCA over the vertex displacements, where we only keep the portion of the components to explain sufficient amount of the dataset variance. Table \ref{tab:bases_dimensions} summarizes the final number of components in each region. The topology of GNM contains $\ngnmverts = 17\,821$ vertices (including the outer skin, teeth, tongue and eyeballs) and the skeletal structure consists of $\ngnmjoints = 4$ joints. The template head mesh $\gnmverttempl$ is placed in the global coordinate system with neck aligned upright along the positive Y-axis, while the face looks along the positive Z-axis.

\begin{table}[!htp]\centering\small
\caption{The dimensions of the individual regions of the identity basis $\idbasis$ and the expression basis $\exprbasis$.}\label{tab:bases_dimensions}
\setlength{\aboverulesep}{0pt}
\setlength{\belowrulesep}{0pt}
\renewcommand{\arraystretch}{1.3}
\begin{tabular}{cccc|cccccc}
\toprule
\multicolumn{4}{c|}{\rule[-1.0ex]{0pt}{4.0ex}\textbf{Identity}} & \multicolumn{6}{c}{\textbf{Expression}} \\
\rule[-1.5ex]{0pt}{4.5ex} head &eye &teeth &\textbf{TOTAL} &left eye &right eye &lower face &tongue &pupil &\textbf{TOTAL} \\ \hline
$170$ &$3$ &$80$ &$\mathbf{253}$ &$100$ &$100$ &$150$ &$31$ &$1$ &$\mathbf{382}$ \\
\bottomrule
\end{tabular}
\end{table}
\section{GNM Functionality \& Experiments}
\label{sec:experiments}
In this section, we present a comprehensive empirical evaluation of the GNM model to validate its statistical soundness, generative capabilities, and performance in common downstream use cases such as in-the-wild face reconstruction. 
We begin with an intrinsic evaluation of the GNM model to assess its core representation characteristics. 
Next, we demonstrate GNM's utility as a robust geometric prior for single-view and multi-view image reconstruction. 
Finally, we create a semantic sampler for the GNM coefficient space based on a dual-CVAE architecture. 
This effectively bypasses the non-intuitive nature of raw PCA spaces, enabling smooth and controllable generation across discrete categories of gender, ethnicity, and twenty distinct action-driven expression classes.
Through these combined benchmarks, we demonstrate that GNM achieves SotA representation and reconstruction performance.

\subsection{Intrinsic Evaluation of GNM}
\label{subsec:intrinsiceval}
Following standard evaluation paradigms in the 3DMM literature \cite{li2017learning, ploumpis2020towards}, our intrinsic evaluation assesses the statistical validity and expressivity of the GNM model. 
A good parametric model must balance \emph{generalization}—its ability to represent diverse face shapes—with \emph{specificity}—its ability to restrict outputs to the plausible manifold of human faces. 
We evaluate the GNM model across these metrics against FLAME \cite{li2017learning}, a widely adopted SotA 3DMM. 
FLAME comes in two primary variants. The first variant models the jaw using a linear shape basis similar to GNM; we refer to this variant as FLAME (w/o Jaw). 
The second, more commonly used variant models the articulation of the jaw through a dedicated jaw joint using LBS; we refer to this variant simply as FLAME. 
Beyond parameterization, the FLAME and GNM models also differ topologically. 
For example, FLAME does not include the inner mouth cavity (teeth, tongue, gums) and captures less of the torso than GNM.
Therefore, to ensure a fair comparison between both models, we restrict our quantitative evaluation to a common facial region, as shown in Figure~\ref{fig:qual_3d_eval}.

\subsubsection{Evaluation dataset}
\label{subsec:evaldataset}
We evaluate both GNM and FLAME on a held-out set of 15\,000 high-resolution 3D scans that were excluded from the training phase of both models. 
Our test scans span a wide variety of identities, facial expressions, and demographic categories (see Figure~\ref{fig:hb_stats}). 
These test scans are registered to the GNM mesh topology using the approach presented in Section \ref{sec:face_registration}.

\subsubsection{Generalization to Novel Shapes}
\label{subsec:generalization}
Generalization measures a model's capacity to represent novel human head shapes. 
To compare the representative capacities of GNM and FLAME while remaining agnostic to the vertex resolution and topology of the 3DMM, we report the scan-to-mesh distance (mm) between the raw ground truth scan and the within-model template mesh to quantify how accurately each model approximates a target scan.

We take a two-stage optimization approach to fit GNM and FLAME to a target scan. 
Fitting here refers to optimizing the parameters of a 3DMM—such as the identity and expression coefficients, joint rotations, and root translation—to obtain a shape that closely approximates the target scan. 
In the first stage of the fitting, we optimize the parameters of the GNM and FLAME models to approximate the registered mesh of the target scan. 
The resulting model parameters serve as an initialization for the second stage of fitting, where the model parameters are further refined using iterative closest point (ICP) constraints to the target scan.
\paragraph{Stage 1: Fitting to the registration mesh.} As the target registrations are provided in the GNM topology, aligning the GNM model to the registration is trivial, as the vertices of the model and the registered shape are already in correspondence. 
However, to align the FLAME model with the registered mesh, we require dense correspondences between the FLAME and GNM meshes to guide the fitting. 

To achieve this, we manually annotate a sparse set of vertices on the FLAME and GNM template meshes that correspond to salient regions such as the corners of the eyes, nose tip, mouth contours, etc. 
Using these sparse correspondences, we estimate a rigid transformation and an isotropic scale that roughly align the FLAME template mesh to the GNM template mesh.
We then non-rigidly  \cite{besl1992method, beeler2011high} deform the vertices of the FLAME template mesh to closely register and align with the shape of the GNM template mesh. 
During this non-rigid deformation step, we restrict the correspondence search to a hand-painted facial mask that excludes peripheral and internal structures of the meshes, such as the inner mouth, ears, nostrils, and eyeballs, preventing deformations based on false vertex associations. 
We run the non-rigid deformation for 10 iterations, regularized by a Laplacian term \cite{Sorkine2004} to prevent geometric artifacts on the aligned FLAME mesh.
This process results in a mesh in the FLAME topology that approximates the GNM template shape very closely, with an average distance of $<0.02$ mm.
This allows us to compute a dense barycentric mapping between FLAME and GNM, which is needed for our initial fitting of FLAME to the registration of a target scan.
We compute this mapping once in a pre-processing step and use it for the rest of our evaluation.

After computing this mapping, we optimize the parameters of both GNM and FLAME to approximate a given registration mesh using an L2 loss to obtain our initial set of model parameters. 
For this purpose, we optimize the model parameters using the Adam optimizer \cite{kingma2015adam} for 5\,000 steps with a learning rate of $1 \times 10^{-3}$.
\paragraph{Stage 2: Fitting to the target scan.}
While the first stage provides a good initial alignment, using these model parameters directly for quantitative evaluation could unfairly bias the results against FLAME, as minor errors introduced during the vertex mapping computation could skew the metrics. 
Therefore, in a second step, we refine the initial alignment by directly incorporating the ground truth scan.
Specifically, we compute closest point-to-surface constraints between the initially aligned GNM/FLAME meshes and the target scan. These constraints are used to further refine the model parameters to closely follow the true scan surface. This refinement step also uses the Adam optimizer \cite{kingma2015adam} for 5\,000 steps with a learning rate of $1 \times 10^{-3}$. To allow both GNM and FLAME to be maximally expressive, we apply no regularization to the model parameters during this stage.

\begin{figure}[t]
    \centering
    \includegraphics[width=1.0\textwidth]{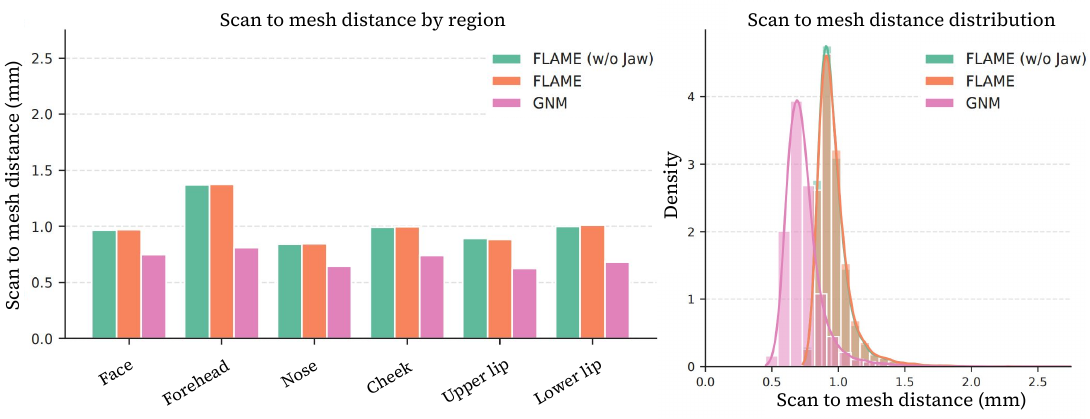}  
    \caption{Reconstruction Error Analysis. On the left, we provide a region-wise breakdown of scan-to-mesh distances (in millimeters) across specific facial areas. On the right, we plot the overall density distribution of these reconstruction errors in the face region. GNM consistently achieves lower reconstruction errors across all individual facial regions and demonstrates a tighter, lower-error overall distribution compared to both FLAME baselines.}
	\label{fig:quantitative_eval}
\end{figure}

\begin{figure}[t]
    \centering
    \includegraphics[width=1.0\textwidth]{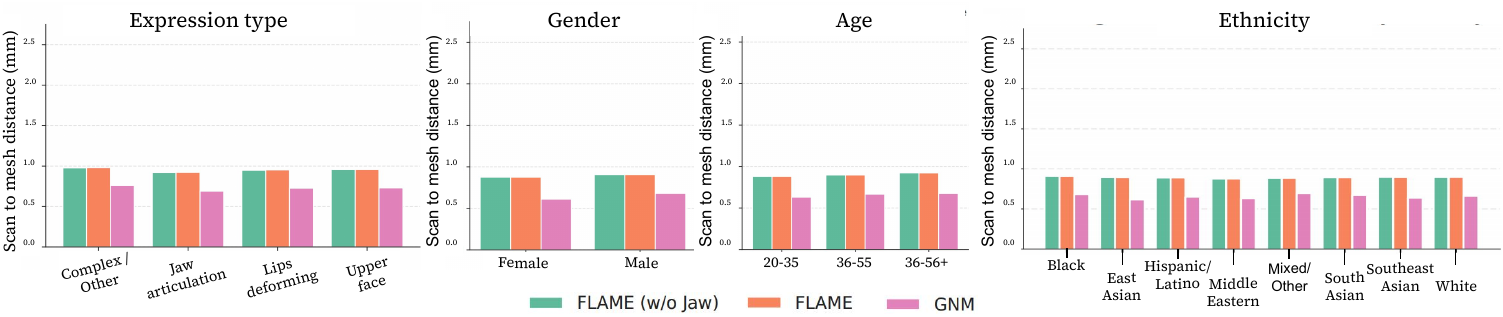}  
    \caption{\textbf{Reconstruction error across demographic and expression subgroups.} Performance is evaluated across four distinct categories: expression type, gender, age, and ethnicity. The GNM model consistently achieves lower reconstruction errors across all evaluated subgroups, demonstrating improved generalization and robustness across diverse demographic populations and facial articulations compared to the FLAME baseline.}
	\label{fig:errorbycat}
\end{figure}

\begin{figure}
    \centering
    \includegraphics[width=\textwidth]{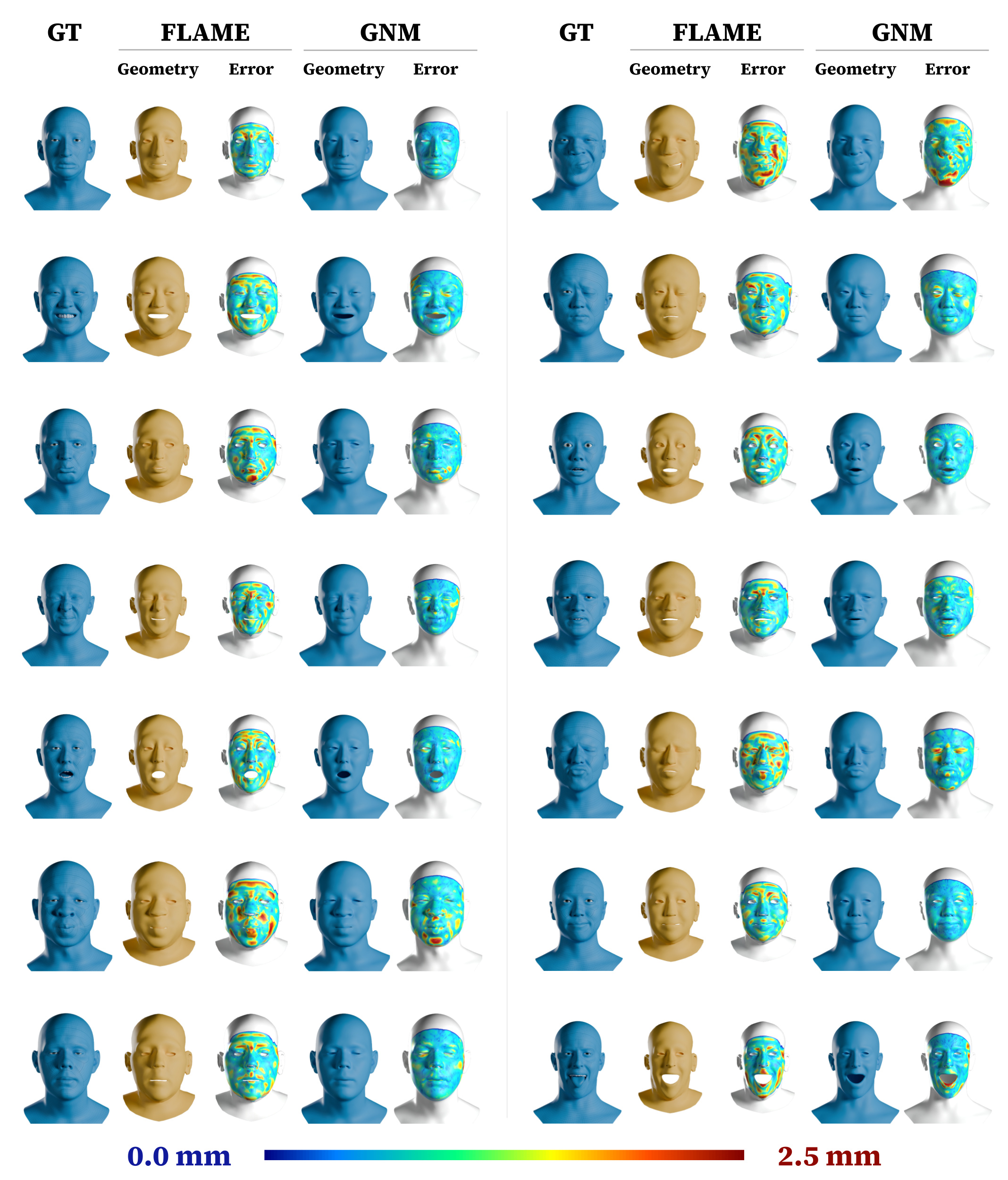}
    \caption{Visual evaluation of the generalization/reconstruction capabilities of FLAME and GNM on unseen ground-truth (GT) scans. GNM achieves noticeably tighter surface alignment (predominantly blue/green) across a diverse range of complex and asymmetrical expressions. Most notably, GNM successfully captures the lower face region during open-mouth/extreme expressions, eliminating the substantial topological errors (visible as red hotspots).}
    \label{fig:qual_3d_eval}
\end{figure}

\paragraph{Quantitative evaluation.}
We repeat this two-stage fitting process independently for each of the 15\,000 scans in our test set.
Table~\ref{tab:flame_vs_gnm} reports the average scan-to-mesh distance for GNM, standard FLAME, and FLAME (w/o Jaw). 
Figure~\ref{fig:quantitative_eval} provides a region-wise breakdown of these errors alongside the overall error distributions. 
A deeper category-wise breakdown of these errors—stratified by expression type, gender, age, and ethnicity—is detailed in Figure~\ref{fig:errorbycat}. 
Finally, in Figure~\ref{fig:qual_3d_eval}, we present qualitative examples, highlighting the spatial distribution of the reconstruction errors via color-coded error maps for both the ground truth and the model approximations.
GNM shows meaningful improvements over FLAME across all metrics and categories, highlighting its potential to be a robust replacement for the FLAME model.

\begin{table}[ht]
\centering
\begin{tabular}{lccc}
\hline
Method & Mean (mm) & Median (mm) & Std Dev (mm) \\
\hline
FLAME (w/o Jaw) \cite{li2017learning} & 0.968 & 0.779 & 0.729 \\
FLAME \cite{li2017learning} & 0.971 & 0.780 & 0.734 \\
\textbf{GNM} & \textbf{0.748} & \textbf{0.623} & \textbf{0.529} \\
\hline
\end{tabular}
\caption{Comparison of scan-to-mesh distances (in millimeters) on a held-out set of 15,000 test scans. GNM demonstrates superior reconstruction accuracy, achieving lower mean, median, and standard deviation errors compared to both FLAME variants.}
\label{tab:flame_vs_gnm}
\end{table}

\begin{figure}[t]
    \centering
    \includegraphics[width=\textwidth]{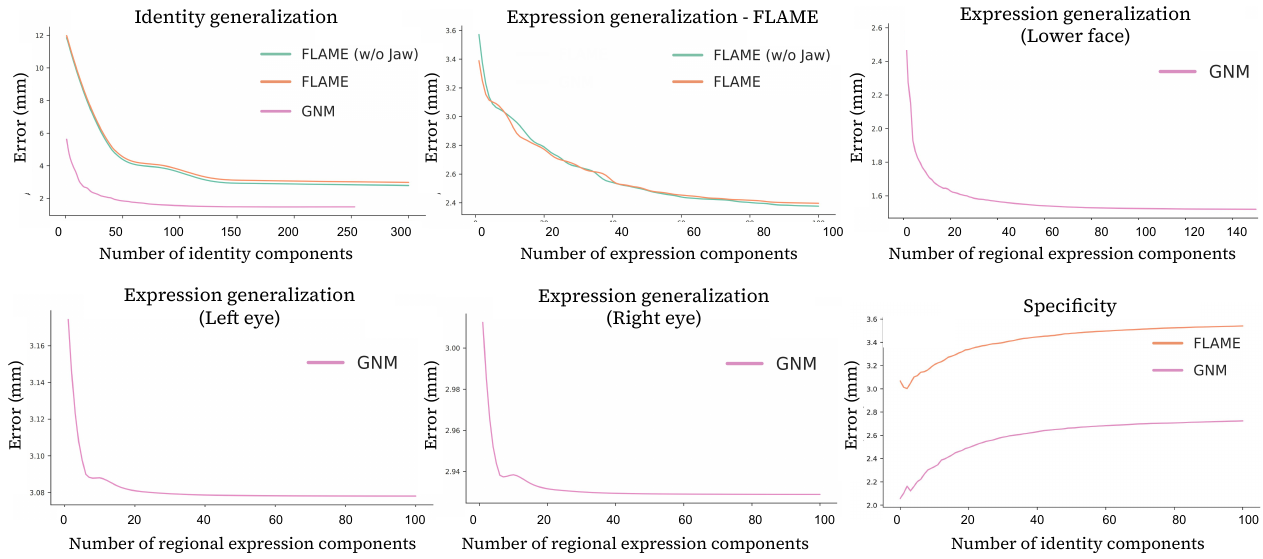}  
    \caption{Intrinsic evaluation of identity and expression spaces. Top row: (Left) Generalization metric of the proposed GNM model against the FLAME model baselines. (Middle) provides the global expression generalization of the FLAME model as a reference. (Right) Lower-face expression generalization for the GNM model. Bottom row: (Left and middle) Left eye and right eye generalization of the GNM model. (Right) Specificity of the GNM and FLAME models. Lower error indicates better performance across all metrics.}
	\label{fig:intrinsic}
\end{figure}

In Figure~\ref{fig:intrinsic}, we also plot the generalization curves for the GNM model's identity and expression bases, that describe how the representative power of the GNM model varies as we vary the number of identity and expression components.
The plots in Figure~\ref{fig:intrinsic} are computed on our evaluation dataset of 15\,000 test scans, and report the scan-to-mesh distance (mm). 
For facial identity (top left) of Figure~\ref{fig:intrinsic}, we see that while both GNM and FLAME's expressivity improve as more components are added, the GNM model achieves a lower error faster, indicating a higher degree of compactness.  
For facial expression, we present the region-wise generalization curve of  FLAME and GNM model in Figure~\ref{fig:intrinsic}. 
Since GNM is a regional expression model, we report the expression generalization curve for each region (left eye, right eye and lower face) independently.
While the expression bases for the left and right eyes are the same (only mirrored), we observe a small difference in the generalization metric for the two regions. We note that this likely stems from the asymmetries in the data distribution of our finite evaluation set, and does not reflect the model itself.
In Figure~\ref{fig:intrinsic}, (top right) we also provide the expression generalization of the global FLAME expression basis for reference.  
Across both facial identity and expression, GNM achieves a better generalization score than FLAME.

\subsubsection{Specificity of GNM}
Specificity evaluates the biological and physical plausibility of the shapes generated by the parameter space of a 3DMM.
A well-behaved 3DMM manifold should mostly allow for the sampling of valid head geometries that remain realistic.
To evaluate the specificity of the GNM model, we randomly sample 2\,000  coefficients $\in~\real^{\nparamsid}$ from a gaussian distribution ($\mathbf{\beta} \sim \mathcal{N}(\mathbf{0}, \mathbf{\Sigma})$), where $\nparamsid$ denotes the number of identity components of the GNM model. 
We evaluate the GNM identity basis using these sampled coefficients to produce neutral head shapes. 
We then compute the minimum surface-to-surface distance between each randomly generated neutral head and its nearest neighbor in our evaluation database:

$$S = \frac{1}{N_s} \sum_{i=1}^{N_s} \min_{\mathbf{M}_{\text{gt}} \in \mathcal{D}} \text{dist}(\mathbf{M}_{\text{syn}, i}, \mathbf{M}_{\text{gt}})$$
where $N_s$ represents the total number of sampled meshes (in this case, 2\,000), $\mathbf{M}_{\text{syn}, i}$ is the $i$-th randomly generated head shape, $\mathcal{D}$ denotes the evaluation database, $\mathbf{M}_{\text{gt}}$ is a ground truth scan within that database, and $\text{dist}(\cdot, \cdot)$ computes the surface-to-surface distance.
We repeat this process for the FLAME model using its identity basis as well. 
In Figure~\ref{fig:intrinsic}, we show the the specificity plot for GNM and FLAME. 
A lower specificity indicates that the sampled identities remain strictly within the boundaries of realistic human anatomy. 
\subsection{3D Face Reconstruction}
A common application of face 3DMMs is 3D reconstruction from single or multi-view images, where the 3DMM serves as a geometric prior to lift 2D visual data to a plausible, animatable, and accurate 3D shape. 
Following SotA approaches in face reconstruction from dense landmarks \cite{wood20223d, hewitt2024look}, we present experiments under three different scenarios to demonstrate GNM's ability to recover plausible facial geometry from images. 
We first provide a brief description of the approach we use to fit GNM to dense landmark constraints.

\paragraph{Fitting GNM to images and videos}
Our approach to fitting GNM to images and videos follows recent optimization-based methods that fit a 3DMM to dense 2D landmarks detected from an image \cite{wood20223d, hewitt2024look}.
Our goal is to optimize the parameters of the GNM model (i.e., the identity and expression coefficients, joint rotations, root translation, and camera intrinsics) to satisfy the dense landmark constraints via re-projection. 
Given an image, we first run a dense landmark detector \cite{chandran2023continuous, chandran2024infinite} to predict approximately 600 landmarks on the face. 
These 2D landmarks serve as the primary data term in our fitting optimization. 
\begin{figure}
\centering
\includegraphics[width=\textwidth]{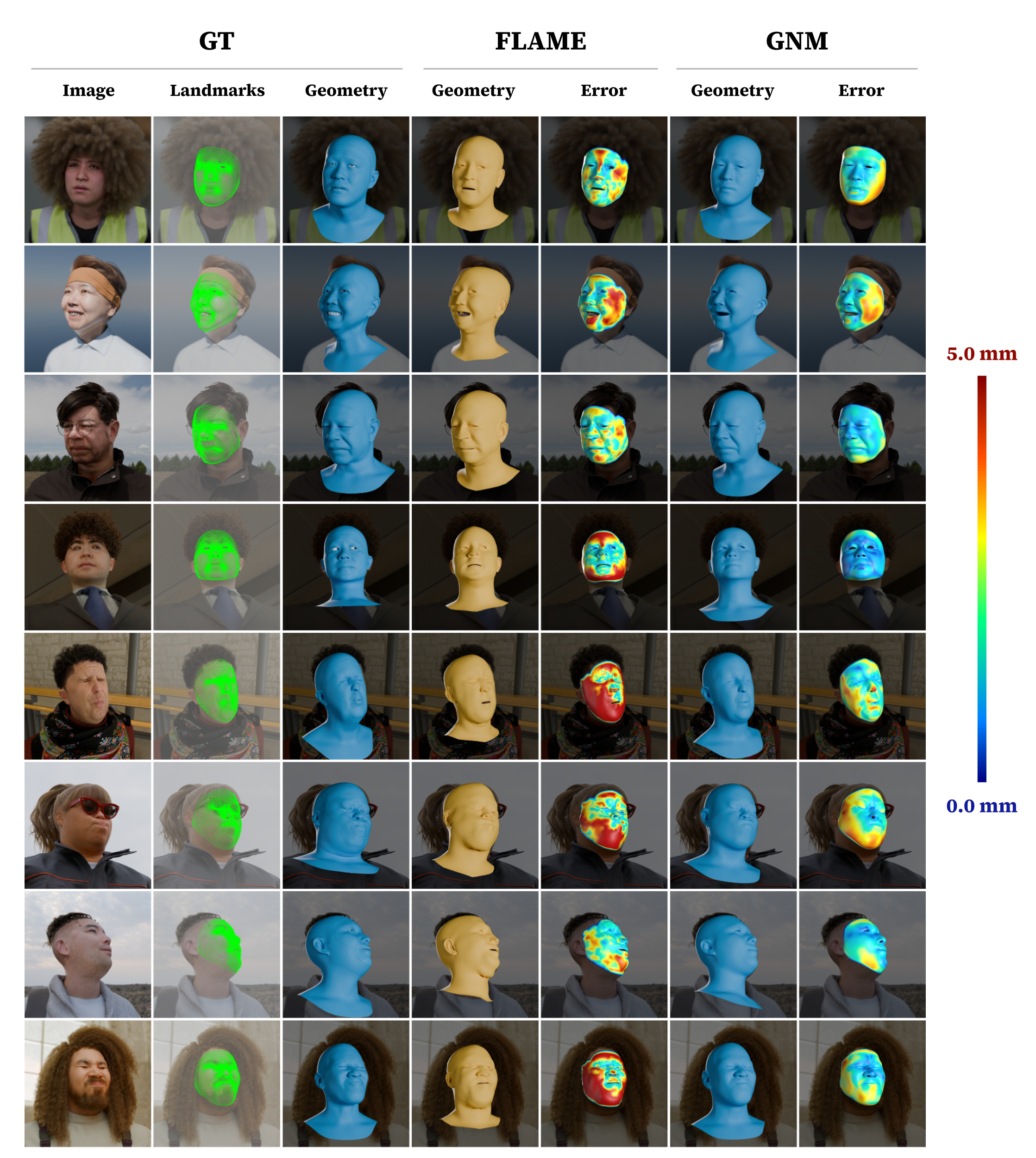}
\caption{\textbf{Qualitative comparison of 3D face reconstruction on synthetic images.} For each synthetic sample, the leftmost panels show the input image, the ground truth (GT) dense surface landmarks (approx. 7\,000 points), and the GT target geometry. The middle panels display the reconstructed geometry and the corresponding color-coded error map for the FLAME model. The rightmost panels show the reconstructed geometry and error map for the GNM model. The error maps visualizes the reconstruction error, illustrating the higher expressivity and tighter fit achieved by GNM.}
\label{fig:synth_qual}
\end{figure}

To prevent geometric artifacts and self-intersections during fitting, we regularize the optimization using a simple L2 prior on the identity and expression coefficients to constrain their magnitude.
We also impose penalties to prevent self-intersections between the skin and internal structures such as the eyeballs and the mouth cavity (teeth, tongue, gums, etc.). 
When optimizing a sequence of images, such as a video, we optionally include a temporal regularization term that encourages the parameters of consecutive frames to remain temporally smooth. 
In summary, our fitting optimization is designed to minimize the weighted sum of four energy terms:
\begin{align}
E_{\text{total}} = w_{\text{lan}} E_{\text{lan}} + w_{\text{prior}} E_{\text{prior}} + w_{\text{anat}} E_{\text{anat}} + w_{\text{temp}} E_{\text{temp}},
\end{align}
where $E_{\text{lan}}$ is the dense landmark re-projection error, $E_{\text{prior}}$ is the L2 regularization on the identity and expression coefficients, $E_{\text{anat}}$ is the anatomical penalty preventing self-intersections, and $E_{\text{temp}}$ is the temporal smoothing term for video sequences. The scalar weights $w_{\text{lan}}$, $w_{\text{prior}}$, $w_{\text{anat}}$, and $w_{\text{temp}}$ determine the relative contribution of each corresponding energy term to the total objective function. For single-frame optimizations, we set $w_{\text{temp}}$ to 0.0.
A detailed explanation of each of these terms and their weighting strategies can be found in \cite{wood20223d}.

\subsubsection{Fitting Comparisons on Single-View Synthetic Data}
To showcase GNM's capacity as a robust shape prior for single-view face reconstruction, we fit the model to a diverse collection of 2\,000 synthetic images, rendered with varying facial identities, expressions, and head poses. 
We selected synthetic data as a fair evaluation baseline for this experiment because it eliminates the uncertainty associated with 2D landmark detection, allowing us to use ground-truth 2D landmarks as the data term. 
We use a highly dense set of 7\,000 ground-truth surface landmarks in the face region to fit both the GNM and FLAME models, applying an identical optimization pipeline to both. 
In Table~\ref{tab:flame_vs_gnm_synth}, we report the average reconstruction metrics across all 2\,000 samples. 
Figure~\ref{fig:synth_qual} provides qualitative examples of fitting FLAME and GNM to these dense landmarks on synthetic images. 
The resulting fits clearly highlight the superior expressivity and geometric fidelity of GNM compared to FLAME.

\begin{table}[ht]
\centering
\begin{tabular}{lccc}
\hline
Method & Mean (mm) & Median (mm) & Std Dev (mm) \\
\hline
FLAME \cite{li2017learning} & 2.172 & 2.086 & \textbf{0.458} \\
\textbf{GNM} & \textbf{1.683} & \textbf{1.589} & 0.591 \\
\hline
\end{tabular}
\caption{\textbf{Quantitative reconstruction evaluation on synthetic data.} We report the mean, median, and standard deviation of the reconstruction error (scan-to-mesh distance in mm) evaluated across 2\,000 single-view synthetic images. Both models are fitted using an identical optimization pipeline driven by 7\,000 GT dense 2D landmarks. GNM achieves a substantially lower mean and median error compared to FLAME, demonstrating its superior capacity as a robust 3D shape prior.}
\label{tab:flame_vs_gnm_synth}
\end{table}

\subsubsection{Fitting to Multi-View Captures}
\begin{figure}
    \centering
    \includegraphics[width=1\linewidth]{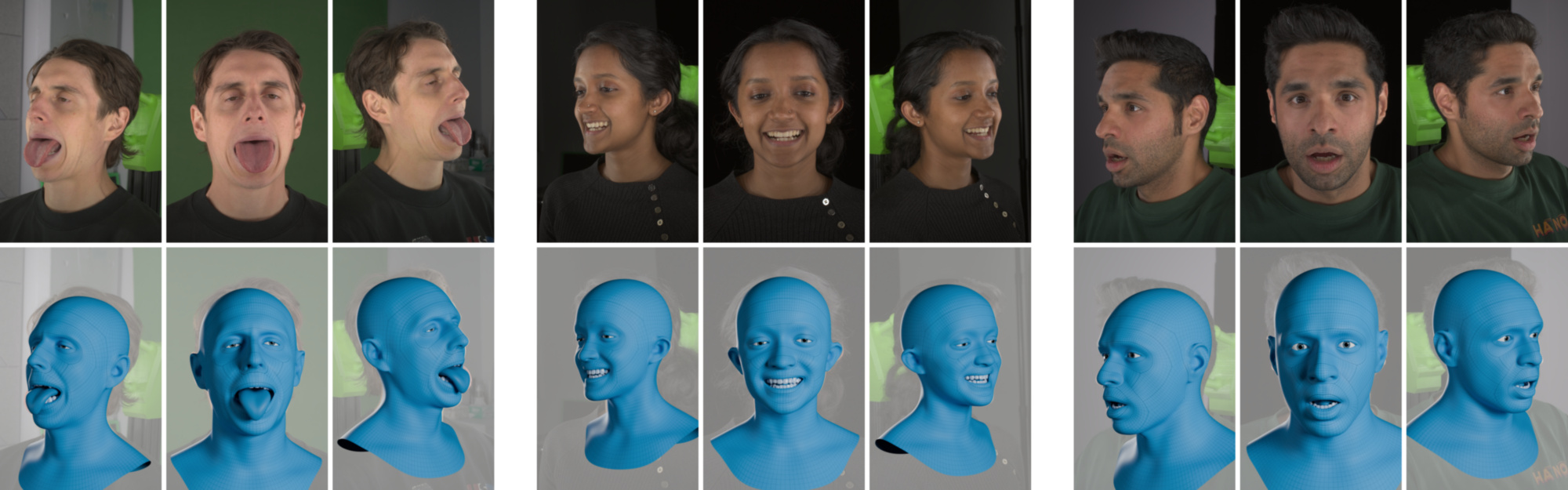} \\[1.5em]

    \includegraphics[width=1\linewidth]{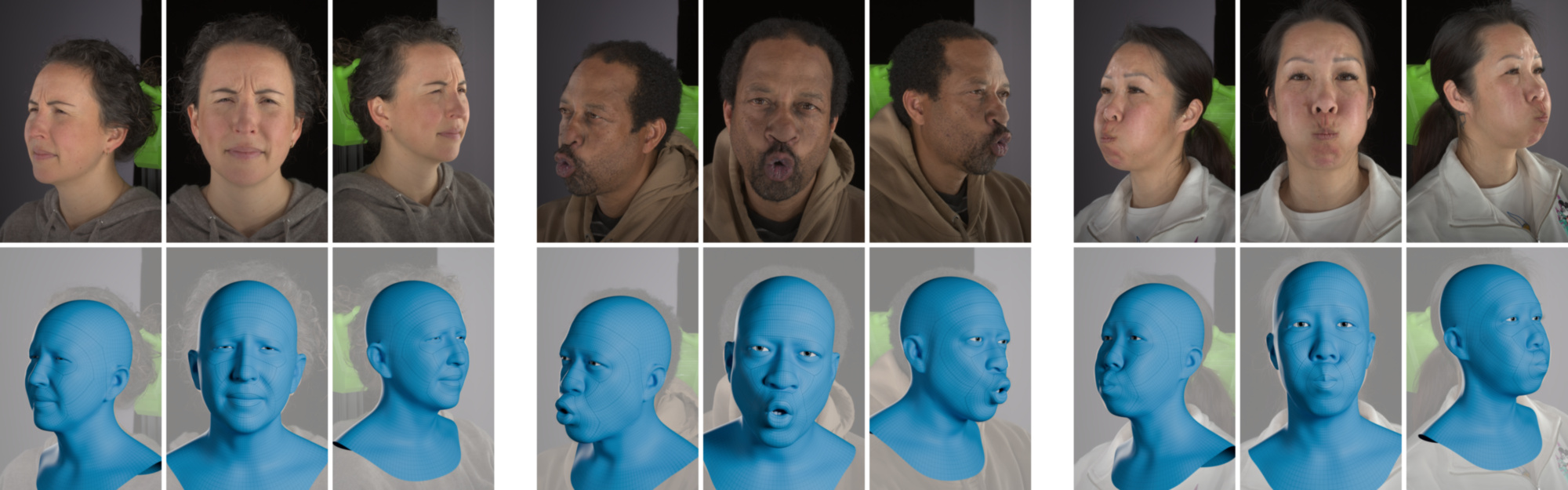} \\[1.5em]

    \includegraphics[width=1\linewidth]{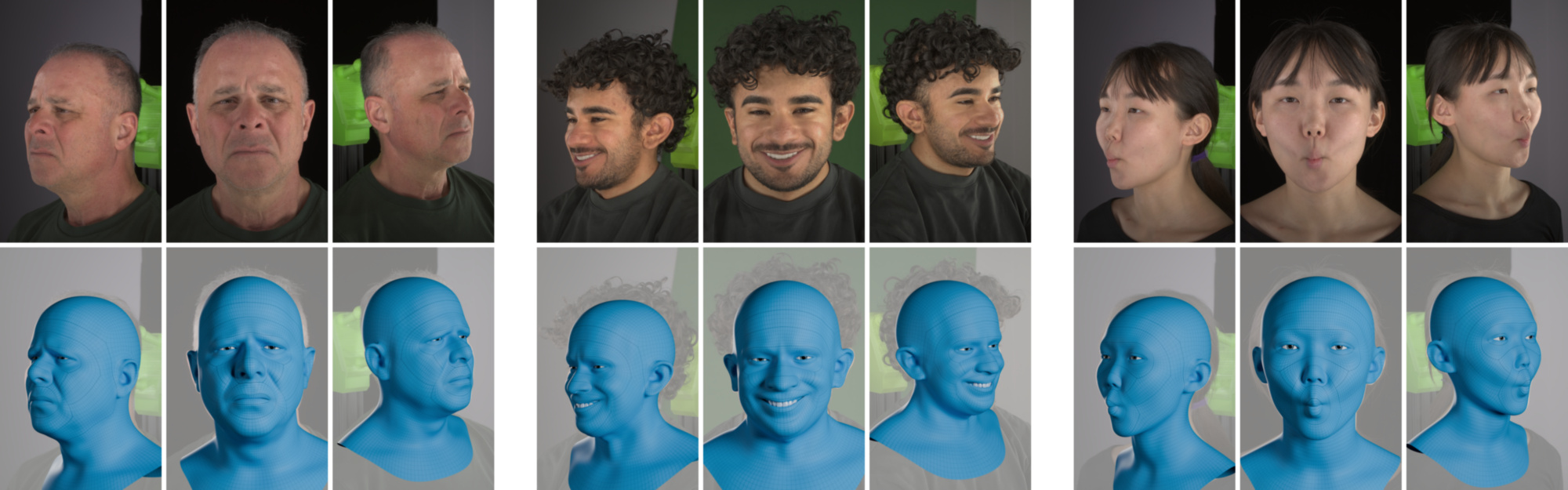} \\[1.5em]
    
    \caption{Multi-view reconstruction images. Top row: Synchronized multi-view input images of extreme expressions. Bottom row: Corresponding GNM fittings. The GNM model is able to recover highly accurate facial identity and expression, and handle complex deformations of both the facial skin and the internal oral cavity, when fit to multiview images.}
    \label{fig:hb_registrations_multiview}
\end{figure}

The second scenario expands our evaluation to fitting GNM to synchronized multi-view videos captured in a controlled studio environment. Here, we slightly modify our optimization pipeline to solve for a single, shared set of identity coefficients across all frames while continuing to solve for per-frame expression coefficients, joint rotations, and root translations. Detailed reconstructions are shown in Figure~\ref{fig:hb_registrations_multiview}, demonstrating our pipeline's ability to handle multi-view inputs with high accuracy. All results depicted in Figure~\ref{fig:hb_registrations_multiview} lie strictly within the shape space of the GNM model and contain no out-of-model deformations.

\subsubsection{Fitting to In-the-Wild Images}

\begin{figure}
    \centering
    \includegraphics[width=\textwidth]{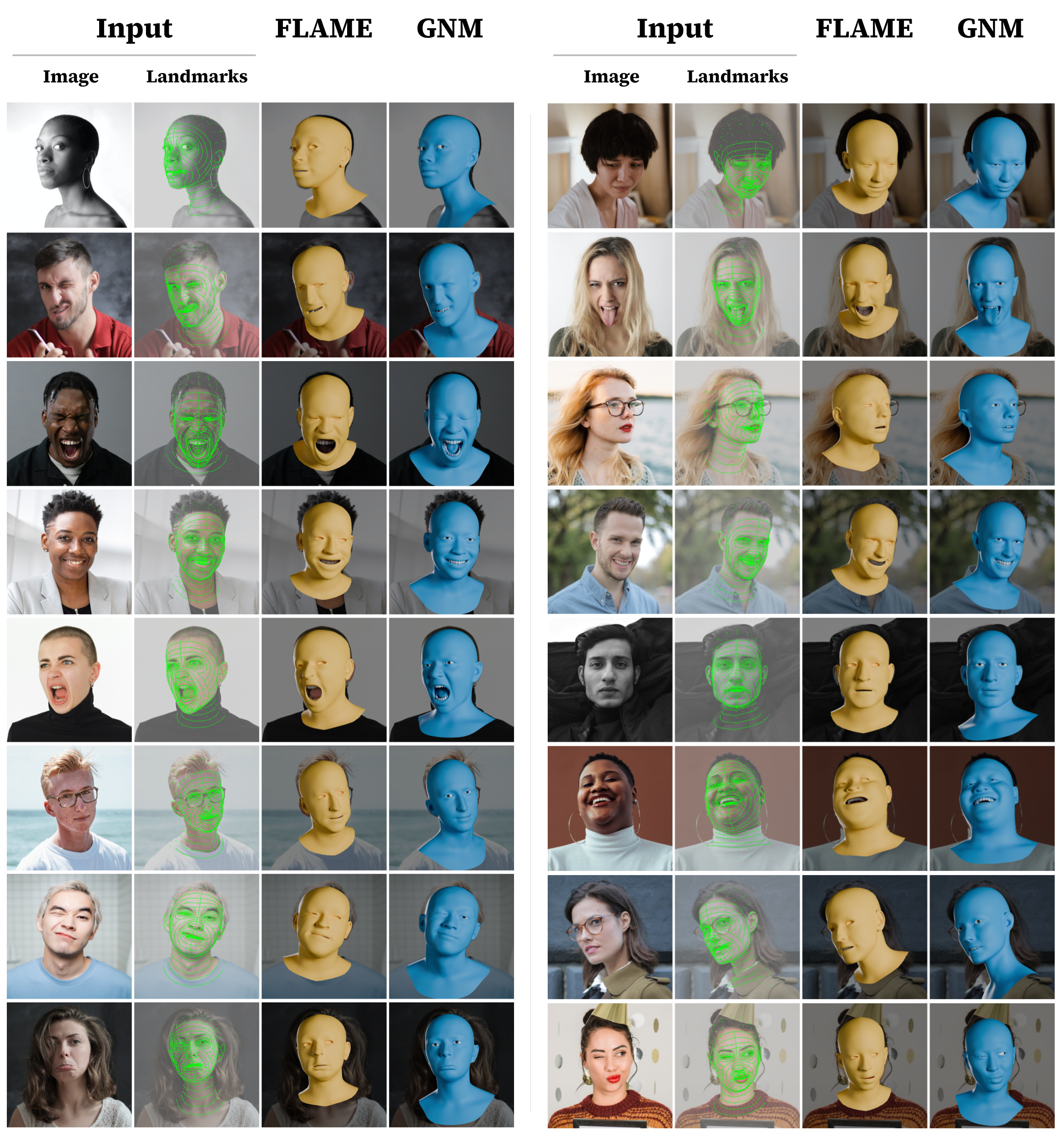}
    \caption{Single-view 3D face reconstruction on in-the-wild (ITW) images. Qualitative comparison of fitting FLAME and GNM to unconstrained single-view images under variable lighting conditions. For each sample block, we show (from left to right): the ground-truth input image, the predicted dense 2D landmarks used to drive the fitting, the FLAME baseline reconstruction, and the proposed GNM reconstruction. GNM demonstrates superior expressivity in capturing extreme, non-linear facial articulations, such as jaw openings, forward tongue expressions and asymmetrical ocular movements.}
    \label{fig:itw_qual}
\end{figure}

\begin{figure}
    \centering
    \includegraphics[width=\textwidth]{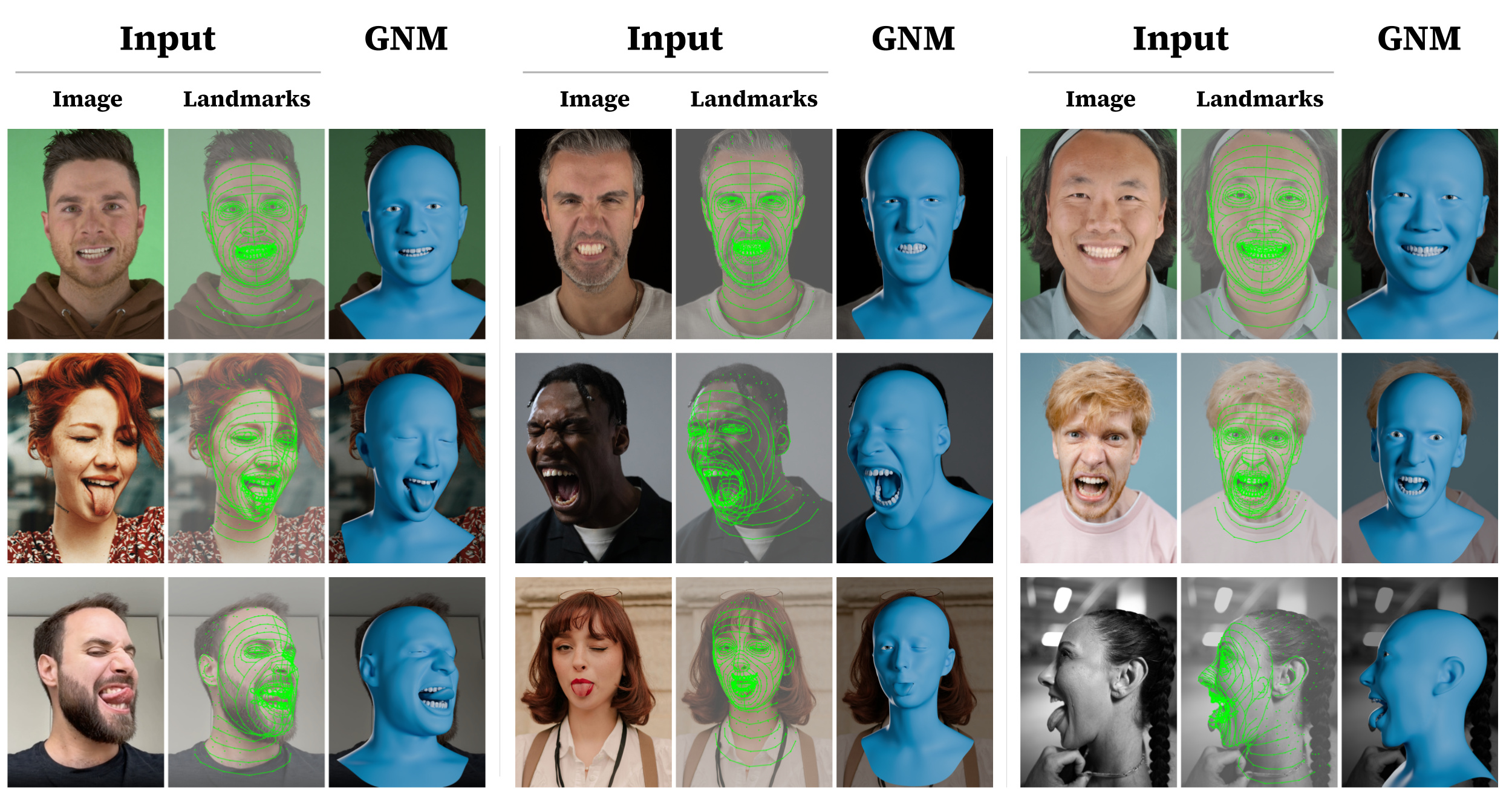}
    \caption{Qualitative evaluation of GNM fitted to extreme, in-the-wild facial expressions. For each example, we display the ground-truth (GT) image, the predicted dense 2D landmarks used to drive the fitting, and the resulting GNM geometry. By explicitly modeling both the visible facial skin and the underlying oral anatomy, GNM successfully recovers fine geometric details; such as individual teeth alignment and extreme tongue expressions, achieving highly accurate reconstructions.}
    \label{fig:itw_qual_mouth}
\end{figure}

Our final scenario evaluates single-view 3D face reconstruction under unconstrained, in-the-wild (ITW) conditions. 
These images feature variable lighting, diverse camera parameters, and extreme facial expressions, posing a significant challenge for robust 3DMM fitting. 
Exemplar reconstructions in these unconstrained settings are shown in Figure~\ref{fig:itw_qual}.

A crucial advantage of the GNM model in these ITW scenarios is its comprehensive modeling of the inner mouth cavity, including the teeth, tongue, and gums which are absent in FLAME. 
By explicitly anchoring the optimization to the detected 2D landmarks for these regions, GNM is able to accurately recover peri-oral deformations, such as extreme jaw openings and tongue expressions. 
As shown in the close-up reconstructions in Figure~\ref{fig:itw_qual_mouth}, GNM successfully recovers fine geometric details, down to individual teeth and extreme tongue poses, leading to overall fits with significantly better structural fidelity than FLAME.
Ultimately, by modeling both the visible facial skin and the underlying oral anatomy, GNM achieves highly detailed, accurate, and anatomically plausible geometries that current models cannot match.
\subsection{Semantically Sampling GNM}
While linear bases constructed via PCA (such as the identity basis $\idbasis$ and expression basis $\exprbasis$) span the anatomical and dynamic variations of the human head, their raw coefficients $\paramsid$ and $\paramsexpr$ are inherently statistical and lack intuitive, localized semantic meaning. A random walk in the raw PCA subspace often leads to global, coupled deformations that alter multiple facial characteristics simultaneously. To enable precise, interpretable generation and targeted attribute editing required by modern animation rigs and conditioning interfaces, we introduce a \emph{Semantic Sampler} which has the ability to generate plausible and human interpretable expressions and identities.

The GNM Semantic Sampler leverages a dual Conditional Variational Autoencoder (CVAE) architecture \cite{kingma2013auto, sohn2015learning} to establish a continuous, differentiable mapping layer. It translates a high-level semantic control vector  $\mathbf{\gamma} \in \mathbb{R}^{D}$ into decoupled GNM identity and expression coefficients: 
\begin{align}
\paramsid = f_{\text{id}}(\mathbf{z}_{\text{id}}, \mathbf{c}_{\text{id}}), \quad \paramsexpr = f_{\text{exp}}(\mathbf{z}_{\text{exp}}, \mathbf{c}_{\text{exp}})
\end{align}

where $\mathbf{c}_{\text{id}}$ and $\mathbf{c}_{\text{exp}}$ represent discrete, one-hot encoded (OHE) conditional category vectors, while $\mathbf{z}_{\text{id}}$ and $\mathbf{z}_{\text{exp}}$ are stochastic latent vectors sampled from a standard normal distribution $\mathcal{N}(\mathbf{0}, \mathbf{I})$ that capture the natural variation within those conditioned classes.

\subsubsection{Network Architecuture and Layer Specifications}

The identity and expression samplers are modeled as standard symmetric CVAEs with encoders and decoders modeled by MLPs with ReLU activations, except for the last layers which are linear. We concatenate the conditioning signal both to the encoder and decoder inputs. Formally, the input to the identity encoder $\mathbf{x}_{\text{id}} = \transpose{\left[ \transpose{\paramsid} \; \transpose{\mathbf{c}_{\text{id}}} \right]}$, while the input to the identity decoder $\mathbf{x}_{\text{id}}' = \transpose{\left[ \transpose{\mathbf{z}_{\text{id}}} \; \transpose{\mathbf{c}_{\text{id}}} \right]}$. Similarly for the decoder, $\mathbf{x}_{\text{exp}} = \transpose{\left[ \transpose{\paramsexpr} \; \transpose{\mathbf{c}_{\text{exp}}} \right]}$ and $\mathbf{x}_{\text{exp}}' = \transpose{\left[ \transpose{\mathbf{z}_{\text{exp}}} \; \transpose{\mathbf{c}_{\text{exp}}} \right]}$. The identity sampler's encoder has 3 layers of sizes $256, 128, 64$, while the expression sampler's encoder has 4 layers of sizer $512, 128, 256, 64$. Their decoders mirror the encoders. Both $\mathbf{z}_{\text{id}}$ and $\mathbf{z}_{\text{exp}}$ representing the latent mean $\mathbf{\mu}$ and log-variance $\log(\mathbf{\sigma}^2)$ are $64$-D vectors. We train the models on a dataset of $12K$ samples.

\subsubsection{Identity Conditioning: Gender and Ethnicity}

\begin{figure}
    \centering
    \includegraphics[width=1.0\linewidth]{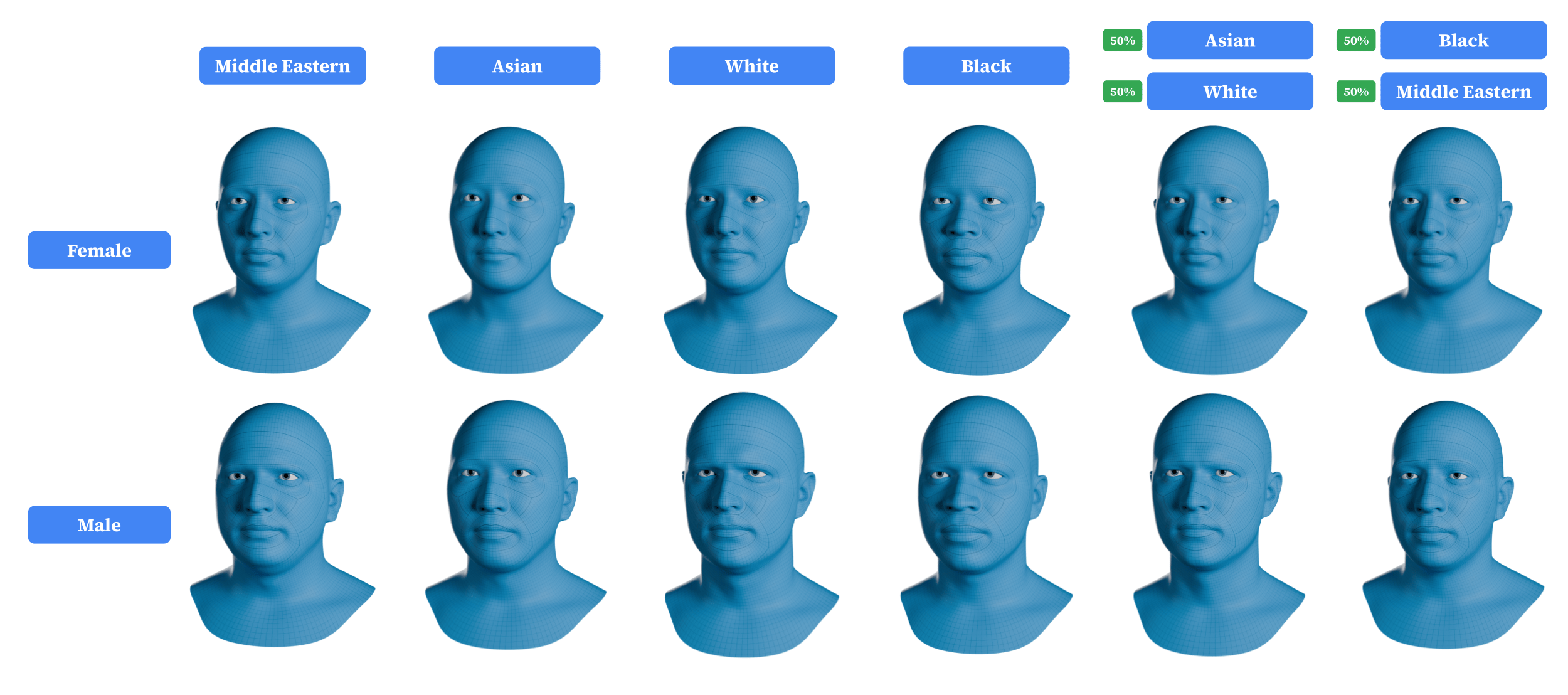}
    \caption{Semantic identity sampling and multi-ethnic interpolation. The GNM Semantic Sampler translates high-level demographic attributes into explicit 3D geometry. These results demonstrate discrete sampling across male and female gender categories and four primary ethnic classes, alongside continuous interpolation between ethnic modalities (far right) achieved through weighted conditional blending.}
    \label{fig:semantic_identity}
\end{figure}

To sample diverse static head structures, the identity conditional vector $\mathbf{c}_{\text{id}}$ is composed of two distinct, horizontally concatenated categorical factors representing gender and ethnicity:
\begin{align*}
\mathbf{c}_{\text{id}} = \left[ \text{OHE}(\text{Gender}) \,\|\, \text{OHE}(\text{Ethnicity}) \right]
\end{align*}

The gender parameter is classified across two classes as male and female, while ethnicity captures demographic variation categorized across four classes: Middle Eastern, Asian, Black and White. By concatenating a 2-dimensional gender OHE vector and a 4-dimensional ethnicity OHE vector, we obtain the unified 6-dimensional conditional vector fed directly to the network. This structured configuration allows for both discrete categorical sampling and smooth, continuous multi-ethnic identity interpolation via weighted category blending, such as generating an identity that is 40\% ethnicity A and 60\% ethnicity B. Some example sampled identities can be seen in Figure~\ref{fig:semantic_identity}.

The identity sampling mechanism is trained on a dataset with specific demographic categorizations. Firstly, we acknowledge that this binary gender classification does not encompass the full spectrum of human gender identities and is not representative of all individuals. Secondly, the datasets used for training group identity variations into four broad ethnic categories. These categories are based on those found in the source 3DMM literature and the available scan data. It is crucial to understand that these groupings are not exhaustive, granular enough, and they do not fully represent the rich diversity of human populations and ancestries worldwide. The use of the two gender and 4 ethnicity categories stems from the limitations of our dataset available for training this type of model. Users should be aware of these limitations and consider the potential implications for fairness, bias, and representation in their specific applications.

\subsubsection{Expression Conditioning: Action-Driven Classes}

\begin{figure}
    \centering
    \includegraphics[width=1.0\linewidth]{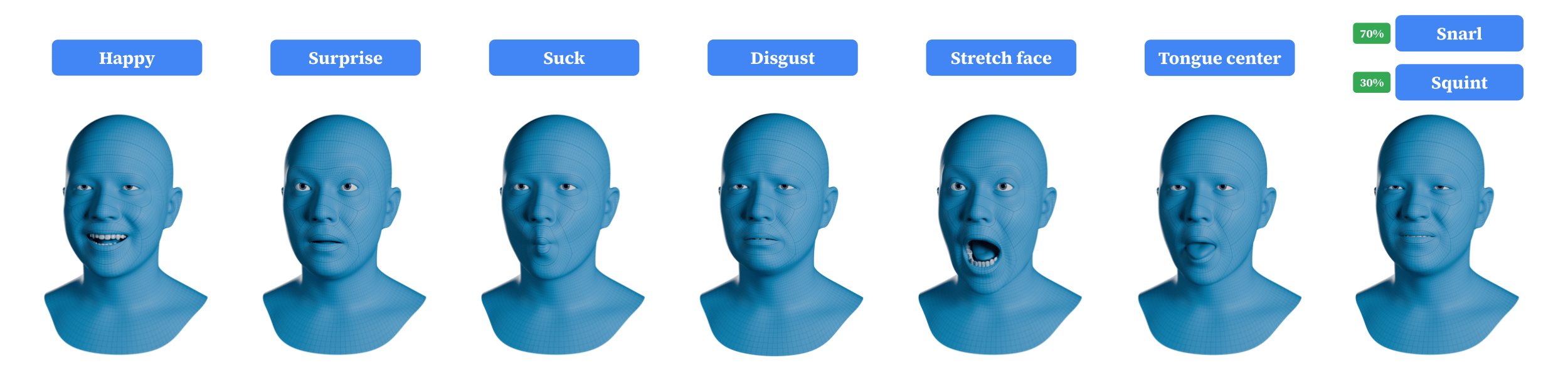}
    \caption{Semantic expression sampling and continuous blending. The GNM Semantic Sampler translates intuitive, action-driven categories into explicit 3D facial deformations. The figure showcases distinct discrete expressions ranging from basic valenced emotions to specific perioral movements alongside a continuously interpolated expression (far right) achieved by applying weighted combinations to the latent conditional vector.}
    \label{fig:semantic_expression}
\end{figure}

For dynamic animations, the expression conditioning vector $\mathbf{c}_{\text{exp}}$ uses a 20-dimensional OHE representation corresponding to 20 distinct, structurally isolated facial movements. These expression step names represent standardized facial action states, spanning primary valenced emotional expressions such as happiness, disgust, surprise, and snarl, alongside detailed perioral and oral movements including a cheek suck, pucker, cheek blow, funneler, lips rolling in, and tongue centering. Additionally, the classes capture structural deformation and tension via compressing and stretching the face, platysma activations, and asymmetrical or highly localized movements such as single eye winking, squinting, sideways mouth motion, wide smile or pulling mouth corners. Similar to the identity model, the expression sampler supports continuous blending. Downstream digital content creation pipelines can generate highly complex, nuanced expressions by applying real-world weight mappings—such as blending happy at 70\% and surprise at 30\% — to yield physically plausible, non-penetrating facial geometries. Some example sampled expressions can be seen in Figure~\ref{fig:semantic_expression}.

\subsubsection{Training Objectives and Latent Space Optimization}

To train the CVAEs to map consistently, we define a joint loss function that balances faithful reconstruction of the original parametric vectors with the regularization of the latent distribution. For a given identity vector $\paramsid$ or expression vector $\paramsexpr$, the network parameters are optimized end-to-end using a weighted objective:
\begin{align}
\mathcal{L}_{\text{CVAE}} = \mathcal{L}_{\text{recon}} + w_{\text{KL}}\mathcal{L}_{\text{KL}}.
\end{align}

The reconstruction term $\mathcal{L}_{\text{recon}}$ is formulated as an $\mathcal{L}_1$ loss rather than an $\mathcal{L}_2$ loss to ensure robust alignment under sharp anatomical boundaries and extreme expressions, preventing the smoothing of high-frequency shape details. The regularization term $\mathcal{L}_{\text{KL}}$ represents the Kullback-Leibler divergence between the learned posterior distribution $q(\mathbf{z} \mid \cdot, \mathbf{c})$ and the prior standard normal distribution $p(\mathbf{z}) = \mathcal{N}(\mathbf{0}, \mathbf{I})$.

To prevent the classic "posterior collapse" notorious in deep fully connected CVAEs, where the network ignores the latent vector $\mathbf{z}$ and relies entirely on the conditional vector $\mathbf{c}$, we implement a cyclical KL annealing schedule \cite{fu2019cyclical}. The weight $w_{\text{KL}}$ is gradually warmed up from $0$ to a maximum threshold of $0.05$ over the first $4\,000$ training steps. 

Furthermore, to guarantee that interpolating between discrete conditions (e.g., generating intermediate ethnicities or blending overlapping facial actions) produces anatomically viable outputs, we apply a mixup regularization technique \cite{zhang2017mixup} to the conditional inputs during training. By feeding the decoder convex combinations of random conditional pairs $\lambda \mathbf{c}_A + (1-\lambda)\mathbf{c}_B$ where $\lambda \sim \text{Beta}(0.2, 0.2)$, the Semantic Sampler learns a highly continuous, globally smooth manifold capable of generating realistic, highly localized facial variations without unnatural geometric distortions.

\section{Discussion and Future Work}
\label{sec:conclusions}

 In this report, we introduced a holistic parametric framework (GNM) that fundamentally advances the digital representation of the human head. By unifying external facial characteristics with the internal ones specifically the oral cavity, teeth, tongue, and ocular anatomy; GNM successfully addresses the critical "hollow shell" limitations inherent in traditional 3DMMs. Built upon an extensive database, GNM establishes a rich latent space both for identity and expression domains. Extensive evaluations confirm that GNM outperforms an established SotA model in generalization and reconstruction accuracy across a wide variety of demographic groups and facial expressions. Beyond its high-fidelity shape representation capability, GNM achieves superior interpretability by decoupling the expression basis into distinct spatial regions, notably the left eye, right eye, and lower face regions. This regional formulation, coupled with our proposed dual-CVAE Semantic Sampler, enables users to achieve semantically meaningful control over diverse identities and facial expressions. Furthermore, we showed that pairing GNM with a robust single and multi-view face fitting framework yields high accuracy in-the-wild reconstructions. GNM is uniquely positioned to act as a foundational structural bridge for next-generation AR telepresence, generative AI conditioning and neural rendering applications. Moving forward, our immediate focus is to expand this open-source ecosystem and release publicly available the dense landmarks utilized for our high-fidelity face-tracking detailed in this report, which will empower broader research community to integrate GNM into their custom pipelines.

\bibliography{main}

\begin{thebibliography}{88}
\providecommand{\natexlab}[1]{#1}
\providecommand{\url}[1]{\texttt{#1}}
\expandafter\ifx\csname urlstyle\endcsname\relax
  \providecommand{\doi}[1]{doi: #1}\else
  \providecommand{\doi}{doi: \begingroup \urlstyle{rm}\Url}\fi

\bibitem[Abdelrehim et~al.(2013)Abdelrehim, Farag, Shalaby, and El-Melegy]{abdelrehim20132d}
A.~S. Abdelrehim, A.~A. Farag, A.~M. Shalaby, and M.~T. El-Melegy.
\newblock 2d-pca shape models: Application to 3d reconstruction of the human teeth from a single image.
\newblock In \emph{International MICCAI Workshop on Medical Computer Vision}, pages 44--52. Springer, 2013.

\bibitem[Alexander et~al.(2010)Alexander, Rogers, Lambeth, Chiang, Ma, Wang, and Debevec]{alexander2010digital}
O.~Alexander, M.~Rogers, W.~Lambeth, J.-Y. Chiang, W.-C. Ma, C.-C. Wang, and P.~Debevec.
\newblock The digital emily project: Achieving a photorealistic digital actor.
\newblock \emph{IEEE Computer Graphics and Applications}, 30\penalty0 (4):\penalty0 20--31, 2010.

\bibitem[Aneja et~al.(2024)Aneja, Thies, Dai, and Nie{\ss}ner]{aneja2024facetalk}
S.~Aneja, J.~Thies, A.~Dai, and M.~Nie{\ss}ner.
\newblock Facetalk: Audio-driven motion diffusion for neural parametric head models.
\newblock In \emph{Proceedings of the IEEE/CVF conference on computer vision and pattern recognition}, pages 21263--21273, 2024.

\bibitem[Bednarik et~al.(2024)Bednarik, Wood, Choutas, Bolkart, Wang, Wu, and Beeler]{bednarik2024learning}
J.~Bednarik, E.~Wood, V.~Choutas, T.~Bolkart, D.~Wang, C.~Wu, and T.~Beeler.
\newblock Learning to stabilize faces.
\newblock In \emph{Computer Graphics Forum}, volume~43, 2024.

\bibitem[Beeler and Bradley(2014)]{beeler2014rigid}
T.~Beeler and D.~Bradley.
\newblock Rigid stabilization of facial expressions.
\newblock \emph{ACM Transactions on Graphics (TOG)}, 33\penalty0 (4):\penalty0 1--9, 2014.

\bibitem[Beeler et~al.(2010)Beeler, Bickel, Beardsley, Sumner, and Gross]{beeler2010high}
T.~Beeler, B.~Bickel, P.~Beardsley, B.~Sumner, and M.~Gross.
\newblock High-quality single-shot capture of facial geometry.
\newblock In \emph{ACM Transactions on Graphics}. Association for Computing Machinery (ACM), 2010.

\bibitem[Beeler et~al.(2011)Beeler, Hahn, Bradley, Bickel, Beardsley, Gotsman, Sumner, and Gross]{beeler2011high}
T.~Beeler, F.~Hahn, D.~Bradley, B.~Bickel, P.~A. Beardsley, C.~Gotsman, R.~W. Sumner, and M.~H. Gross.
\newblock High-quality passive facial performance capture using anchor frames.
\newblock \emph{ACM Trans. Graph.}, 30\penalty0 (4):\penalty0 75, 2011.

\bibitem[B{\'e}rard et~al.(2014)B{\'e}rard, Bradley, Nitti, Beeler, and Gross]{berard2014high}
P.~B{\'e}rard, D.~Bradley, M.~Nitti, T.~Beeler, and M.~H. Gross.
\newblock High-quality capture of eyes.
\newblock \emph{ACM Trans. Graph.}, 33:\penalty0 223--1, 2014.

\bibitem[B{\'e}rard et~al.(2016)B{\'e}rard, Bradley, Gross, and Beeler]{berard2016lightweight}
P.~B{\'e}rard, D.~Bradley, M.~Gross, and T.~Beeler.
\newblock Lightweight eye capture using a parametric model.
\newblock \emph{ACM Transactions on Graphics (TOG)}, 35\penalty0 (4):\penalty0 1--12, 2016.

\bibitem[B{\'e}rard et~al.(2019)B{\'e}rard, Bradley, Gross, and Beeler]{berard2019practical}
P.~B{\'e}rard, D.~Bradley, M.~Gross, and T.~Beeler.
\newblock Practical person-specific eye rigging.
\newblock In \emph{Computer Graphics Forum}, volume~38, 2019.

\bibitem[Besl and McKay(1992)]{besl1992method}
P.~J. Besl and N.~D. McKay.
\newblock Method for registration of 3-d shapes.
\newblock In \emph{Sensor fusion IV: control paradigms and data structures}, volume 1611, pages 586--606. Spie, 1992.

\bibitem[Blanz and Vetter(1999)]{blanz1999morphable}
V.~Blanz and T.~Vetter.
\newblock A morphable model for the synthesis of 3d faces.
\newblock In \emph{International Conference on Computer Graphics and Interactive Techniques}, pages 187--194. ACM Press, 1999.
\newblock \doi{10.1145/3596711.3596730}.

\bibitem[Booth et~al.(2016)Booth, Roussos, Zafeiriou, Ponniah, and Dunaway]{booth20163d}
J.~Booth, A.~Roussos, S.~Zafeiriou, A.~Ponniah, and D.~Dunaway.
\newblock A 3d morphable model learnt from 10,000 faces.
\newblock In \emph{Proceedings of the IEEE conference on computer vision and pattern recognition}, pages 5543--5552, 2016.

\bibitem[Buehler et~al.(2024)Buehler, Li, Wood, Helminger, Chen, Shah, Wang, Garbin, Orts-Escolano, Hilliges, Lagun, Riviere, Gotardo, Beeler, Meka, and Sarkar]{buehler2024cafca}
M.~C. Buehler, G.~Li, E.~Wood, L.~Helminger, X.~Chen, T.~Shah, D.~Wang, S.~Garbin, S.~Orts-Escolano, O.~Hilliges, D.~Lagun, J.~Riviere, P.~Gotardo, T.~Beeler, A.~Meka, and K.~Sarkar.
\newblock Cafca: High-quality novel view synthesis of expressive faces from casual few-shot captures.
\newblock In \emph{SIGGRAPH}. ACM, 2024.

\bibitem[Cao et~al.(2014)Cao, Weng, Zhou, Tong, and Zhou]{cao2013facewarehouse}
C.~Cao, Y.~Weng, S.~Zhou, Y.~Tong, and K.~Zhou.
\newblock Facewarehouse: A 3d facial expression database for visual computing.
\newblock \emph{IEEE Transactions on Visualization and Computer Graphics}, 20\penalty0 (3):\penalty0 413--425, 2014.
\newblock \doi{10.1109/TVCG.2013.249}.

\bibitem[Chandran and Zoss(2024)]{chandran2024anatomically}
P.~Chandran and G.~Zoss.
\newblock Anatomically constrained implicit face models.
\newblock In \emph{Proceedings of the IEEE/CVF Conference on Computer Vision and Pattern Recognition}, pages 2220--2229, 2024.

\bibitem[Chandran et~al.(2022)Chandran, Zoss, Gross, Gotardo, and Bradley]{chandran2022shape}
P.~Chandran, G.~Zoss, M.~Gross, P.~Gotardo, and D.~Bradley.
\newblock Shape transformers: Topology-independent 3d shape models using transformers.
\newblock In \emph{Computer Graphics Forum}, volume~41, pages 195--207. Wiley Online Library, 2022.

\bibitem[Chandran et~al.(2023)Chandran, Zoss, Gotardo, and Bradley]{chandran2023continuous}
P.~Chandran, G.~Zoss, P.~Gotardo, and D.~Bradley.
\newblock Continuous landmark detection with 3d queries.
\newblock In \emph{Proceedings of the IEEE/CVF Conference on Computer Vision and Pattern Recognition}, pages 16858--16867, 2023.

\bibitem[Chandran et~al.(2024)Chandran, Zoss, Gotardo, and Bradley]{chandran2024infinite}
P.~Chandran, G.~Zoss, P.~Gotardo, and D.~Bradley.
\newblock Infinite 3d landmarks: Improving continuous 2d facial landmark detection.
\newblock In \emph{Computer Graphics Forum}, volume~43, page e15126. Wiley Online Library, 2024.

\bibitem[Chen et~al.(2024)Chen, Mihajlovic, Wang, Prokudin, and Tang]{chen2024morphable}
X.~Chen, M.~Mihajlovic, S.~Wang, S.~Prokudin, and S.~Tang.
\newblock Morphable diffusion: 3d-consistent diffusion for single-image avatar creation.
\newblock In \emph{Proceedings of the IEEE/CVF Conference on Computer Vision and Pattern Recognition}, pages 10359--10370, 2024.

\bibitem[Chu and Harada(2024)]{chu2024generalizable}
X.~Chu and T.~Harada.
\newblock Generalizable and animatable gaussian head avatar.
\newblock \emph{Advances in Neural Information Processing Systems}, 37:\penalty0 57642--57670, 2024.

\bibitem[Cudeiro et~al.(2019)Cudeiro, Bolkart, Laidlaw, Ranjan, and Black]{cudeiro2019capture}
D.~Cudeiro, T.~Bolkart, C.~Laidlaw, A.~Ranjan, and M.~J. Black.
\newblock Capture, learning, and synthesis of 3d speaking styles.
\newblock In \emph{Proceedings of the IEEE/CVF conference on computer vision and pattern recognition}, pages 10101--10111, 2019.

\bibitem[Dai et~al.(2018)Dai, Pears, and Smith]{dai2018data}
H.~Dai, N.~Pears, and W.~Smith.
\newblock A data-augmented 3d morphable model of the ear.
\newblock In \emph{2018 13th IEEE International Conference on Automatic Face \& Gesture Recognition (FG 2018)}, pages 404--408. IEEE, 2018.

\bibitem[Dan{\v{e}}{\v{c}}ek et~al.(2022)Dan{\v{e}}{\v{c}}ek, Black, and Bolkart]{danecek2022emoca}
R.~Dan{\v{e}}{\v{c}}ek, M.~J. Black, and T.~Bolkart.
\newblock Emoca: Emotion driven monocular face capture and animation.
\newblock In \emph{Proceedings of the IEEE/CVF conference on computer vision and pattern recognition}, pages 20311--20322, 2022.

\bibitem[Danecek et~al.(2025)Danecek, Schmitt, Polikovsky, and Black]{danecek2025supervising}
R.~Danecek, C.~Schmitt, S.~Polikovsky, and M.~J. Black.
\newblock Supervising 3d talking head avatars with analysis-by-audio-synthesis.
\newblock In \emph{Thirteenth International Conference on 3D Vision}, 2025.

\bibitem[Edwards et~al.(2020)Edwards, Landreth, Pop{\l}awski, Malinowski, Watling, Fiume, and Singh]{edwards2020jali}
P.~Edwards, C.~Landreth, M.~Pop{\l}awski, R.~Malinowski, S.~Watling, E.~Fiume, and K.~Singh.
\newblock Jali-driven expressive facial animation and multilingual speech in cyberpunk 2077.
\newblock In \emph{Special Interest Group on Computer Graphics and Interactive Techniques Conference Talks}, pages 1--2, 2020.

\bibitem[Egger et~al.(2020)Egger, Smith, Tewari, Wuhrer, Zollhöfer, Beeler, Bernard, Bolkart, Kortylewski, Romdhani, Theobalt, Blanz, and Vetter]{Egger2020:3DMM}
B.~Egger, W.~Smith, A.~Tewari, S.~Wuhrer, M.~Zollhöfer, T.~Beeler, F.~Bernard, T.~Bolkart, A.~Kortylewski, S.~Romdhani, C.~Theobalt, V.~Blanz, and T.~Vetter.
\newblock {3D} morphable face models--past, present and future.
\newblock \emph{ACM Transactions on Graphics}, 39\penalty0 (5):\penalty0 1--38, 2020.
\newblock \doi{10.1145/3395208}.

\bibitem[{Epic Games}(2026)]{epicgames2026metahuman}
{Epic Games}.
\newblock {Unreal Engine MetaHuman}.
\newblock \url{https://unrealengine.com}, 2026.

\bibitem[Fan et~al.(2022)Fan, Lin, Saito, Wang, and Komura]{fan2022faceformer}
Y.~Fan, Z.~Lin, J.~Saito, W.~Wang, and T.~Komura.
\newblock Faceformer: Speech-driven 3d facial animation with transformers.
\newblock In \emph{Proceedings of the IEEE/CVF conference on computer vision and pattern recognition}, pages 18770--18780, 2022.

\bibitem[Feng et~al.(2021)Feng, Feng, Black, and Bolkart]{feng2021learning}
Y.~Feng, H.~Feng, M.~J. Black, and T.~Bolkart.
\newblock Learning an animatable detailed 3d face model from in-the-wild images.
\newblock \emph{ACM Transactions on Graphics (ToG)}, 40\penalty0 (4):\penalty0 1--13, 2021.

\bibitem[Fu et~al.(2019)Fu, Li, Liu, Gao, Celikyilmaz, and Carin]{fu2019cyclical}
H.~Fu, C.~Li, X.~Liu, J.~Gao, A.~Celikyilmaz, and L.~Carin.
\newblock Cyclical annealing schedule: A simple approach to mitigating kl vanishing.
\newblock In \emph{Proceedings of the 2019 Conference of the North American Chapter of the Association for Computational Linguistics: Human Language Technologies, Volume 1 (Long and Short Papers)}, pages 240--250, 2019.

\bibitem[Giebenhain et~al.(2023)Giebenhain, Kirschstein, Georgopoulos, Rünz, Agapito, and Nießner]{giebenhain2023learning}
S.~Giebenhain, T.~Kirschstein, M.~Georgopoulos, M.~Rünz, L.~Agapito, and M.~Nießner.
\newblock Learning neural parametric head models.
\newblock In \emph{Computer Vision and Pattern Recognition}, pages 21003--21012, 2023.
\newblock \doi{10.1109/CVPR52729.2023.02012}.

\bibitem[Giebenhain et~al.(2024)Giebenhain, Kirschstein, R{\"u}nz, Agapito, and Nie{\ss}ner]{giebenhain2024npga}
S.~Giebenhain, T.~Kirschstein, M.~R{\"u}nz, L.~Agapito, and M.~Nie{\ss}ner.
\newblock Npga: Neural parametric gaussian avatars.
\newblock In \emph{SIGGRAPH Asia 2024 Conference Papers}, pages 1--11, 2024.

\bibitem[Guestrin and Eizenman(2006)]{guestrin2006general}
E.~D. Guestrin and M.~Eizenman.
\newblock General theory of remote gaze estimation using the pupil center and corneal reflections.
\newblock \emph{IEEE Transactions on Biomedical Engineering}, 53\penalty0 (6):\penalty0 1124--1133, 2006.
\newblock \doi{10.1109/TBME.2005.863952}.

\bibitem[Guo et~al.(2019)Guo, Lincoln, Davidson, Busch, Yu, Whalen, Harvey, Orts-Escolano, Pandey, Dourgarian, Danhang, Tkach, Kowdle, Cooper, Dou, Fanello, Fyffe, Rhemann, Taylor, Debevec, and Izadi]{guo2019relightables}
K.~Guo, P.~Lincoln, P.~Davidson, J.~Busch, X.~Yu, M.~Whalen, G.~Harvey, S.~Orts-Escolano, R.~Pandey, J.~Dourgarian, T.~Danhang, A.~Tkach, A.~Kowdle, E.~Cooper, M.~Dou, S.~Fanello, G.~Fyffe, C.~Rhemann, J.~Taylor, P.~Debevec, and S.~Izadi.
\newblock The relightables: Volumetric performance capture of humans with realistic relighting.
\newblock \emph{ACM Transactions on Graphics (ToG)}, 38\penalty0 (6):\penalty0 1--19, 2019.

\bibitem[Harling(2018)]{gdpr2016}
G.~Harling.
\newblock General data protection regulation (gdpr).
\newblock \emph{Official Journal of the European Union}, 2018.

\bibitem[Hewitt et~al.(2024)Hewitt, Saleh, Aliakbarian, Petikam, Rezaeifar, Florentin, Hosenie, Cashman, Valentin, Cosker, and Baltru\v{s}aitis]{hewitt2024look}
C.~Hewitt, F.~Saleh, S.~Aliakbarian, L.~Petikam, S.~Rezaeifar, L.~Florentin, Z.~Hosenie, T.~J. Cashman, J.~Valentin, D.~Cosker, and T.~Baltru\v{s}aitis.
\newblock Look ma, no markers: holistic performance capture without the hassle.
\newblock \emph{ACM Transactions on Graphics (TOG)}, 43\penalty0 (6), 2024.

\bibitem[Hirshberg et~al.(2012)Hirshberg, Loper, Rachlin, and Black]{hirshberg2012coregistration}
D.~A. Hirshberg, M.~Loper, E.~Rachlin, and M.~J. Black.
\newblock Coregistration: Simultaneous alignment and modeling of articulated {3D} shape.
\newblock In \emph{Computer Vision - {ECCV} 2012 - 12th European Conference on Computer Vision, Florence, Italy, October 7-13, 2012, Proceedings, Part {VI}}, volume 7577 of \emph{Lecture Notes in Computer Science}, pages 242--255. Springer, 2012.

\bibitem[Jakob et~al.(2022)Jakob, Speierer, Roussel, Nimier-David, Vicini, Zeltner, Nicolet, Crespo, Leroy, and Zhang]{Mitsuba3}
W.~Jakob, S.~Speierer, N.~Roussel, M.~Nimier-David, D.~Vicini, T.~Zeltner, B.~Nicolet, M.~Crespo, V.~Leroy, and Z.~Zhang.
\newblock Mitsuba 3 renderer.
\newblock \url{https://mitsuba-renderer.org}, 2022.
\newblock Version 3.8.0.

\bibitem[Kerbl et~al.(2023)Kerbl, Kopanas, Leimkuehler, and Drettakis]{kerbl20233d}
B.~Kerbl, G.~Kopanas, T.~Leimkuehler, and G.~Drettakis.
\newblock 3d gaussian splatting for real-time radiance field rendering.
\newblock \emph{ACM Transactions on Graphics}, 42\penalty0 (4):\penalty0 1--14, 2023.
\newblock \doi{10.1145/3592433}.

\bibitem[Kingma and Ba(2015)]{kingma2015adam}
D.~P. Kingma and J.~Ba.
\newblock Adam: A method for stochastic optimization.
\newblock In \emph{International Conference on Learning Representations (ICLR)}, 2015.
\newblock URL \url{https://arxiv.org/abs/1412.6980}.

\bibitem[Kingma and Welling(2013)]{kingma2013auto}
D.~P. Kingma and M.~Welling.
\newblock Auto-encoding variational bayes.
\newblock \emph{arXiv preprint arXiv:1312.6114}, 2013.

\bibitem[Li et~al.(2022)Li, Meka, Mueller, Buehler, Hilliges, and Beeler]{li2022eyenerf}
G.~Li, A.~Meka, F.~Mueller, M.~C. Buehler, O.~Hilliges, and T.~Beeler.
\newblock Eyenerf: a hybrid representation for photorealistic synthesis, animation and relighting of human eyes.
\newblock \emph{ACM Transactions on Graphics (ToG)}, 41\penalty0 (4):\penalty0 1--16, 2022.

\bibitem[Li et~al.(2017)Li, Bolkart, Black, Li, and Romero]{li2017learning}
T.~Li, T.~Bolkart, M.~J. Black, H.~Li, and J.~Romero.
\newblock Learning a model of facial shape and expression from 4d scans.
\newblock \emph{ACM Transactions on Graphics}, 36\penalty0 (6):\penalty0 1--17, 2017.
\newblock \doi{10.1145/3130800.3130813}.

\bibitem[Li et~al.(2018)Li, Aittala, Durand, and Lehtinen]{Li2018}
T.-M. Li, M.~Aittala, F.~Durand, and J.~Lehtinen.
\newblock Differentiable monte carlo ray tracing through edge sampling.
\newblock \emph{ACM Trans. Graph. (Proc. SIGGRAPH Asia)}, 37\penalty0 (6):\penalty0 222:1--222:11, 2018.

\bibitem[Luo and Hancock(2002)]{luo2002iterative}
B.~Luo and E.~R. Hancock.
\newblock Iterative procrustes alignment with the em algorithm.
\newblock \emph{Image and Vision Computing}, 20\penalty0 (5-6):\penalty0 377--396, 2002.

\bibitem[Medina et~al.(2022)Medina, Tome, Stoll, Tiede, Munhall, Hauptmann, and Matthews]{medina2022speech}
S.~Medina, D.~Tome, C.~Stoll, M.~Tiede, K.~Munhall, A.~G. Hauptmann, and I.~Matthews.
\newblock Speech driven tongue animation.
\newblock In \emph{Computer Vision and Pattern Recognition}, pages 20374--20384. IEEE, 2022.
\newblock \doi{10.1109/CVPR52688.2022.01976}.

\bibitem[Mildenhall et~al.(2021)Mildenhall, Srinivasan, Tancik, Barron, Ramamoorthi, and Ng]{mildenhall2021nerf}
B.~Mildenhall, P.~Srinivasan, M.~Tancik, J.~T. Barron, R.~Ramamoorthi, and R.~Ng.
\newblock Nerf: Representing scenes as neural radiance fields for view synthesis.
\newblock \emph{Commun. ACM}, 65\penalty0 (1):\penalty0 405--421, 2021.
\newblock \doi{10.1145/3503250}.

\bibitem[Nicolet et~al.(2021)Nicolet, Jacobson, and Jakob]{Nicolet2021Large}
B.~Nicolet, A.~Jacobson, and W.~Jakob.
\newblock Large steps in inverse rendering of geometry.
\newblock \emph{ACM Transactions on Graphics (Proceedings of SIGGRAPH Asia)}, 40\penalty0 (6), Dec. 2021.
\newblock \doi{10.1145/3478513.3480501}.

\bibitem[O'Sullivan et~al.(2021)O'Sullivan, van~de Lande, Oosting, Papaioannou, Jeelani, Koudstaal, Khonsari, Dunaway, Zafeiriou, and Schievano]{o20213d}
E.~O'Sullivan, L.~S. van~de Lande, A.-J.~C. Oosting, A.~Papaioannou, N.~O. Jeelani, M.~J. Koudstaal, R.~H. Khonsari, D.~J. Dunaway, S.~Zafeiriou, and S.~Schievano.
\newblock The 3d skull 0--4 years: a validated, generative, statistical shape model.
\newblock \emph{Bone reports}, 15:\penalty0 101154, 2021.

\bibitem[O'Sullivan et~al.(2022)O'Sullivan, van~de Lande, El~Ghoul, Koudstaal, Schievano, Khonsari, Dunaway, and Zafeiriou]{o2022growth}
E.~O'Sullivan, L.~S. van~de Lande, K.~El~Ghoul, M.~J. Koudstaal, S.~Schievano, R.~H. Khonsari, D.~J. Dunaway, and S.~Zafeiriou.
\newblock Growth patterns and shape development of the paediatric mandible--a 3d statistical model.
\newblock \emph{Bone reports}, 16:\penalty0 101528, 2022.

\bibitem[Paysan et~al.(2009)Paysan, Knothe, Amberg, Romdhani, and Vetter]{paysan20093d}
P.~Paysan, R.~Knothe, B.~Amberg, S.~Romdhani, and T.~Vetter.
\newblock A 3d face model for pose and illumination invariant face recognition.
\newblock In \emph{2009 sixth IEEE international conference on advanced video and signal based surveillance}, pages 296--301. IEEE, IEEE, 2009.
\newblock \doi{10.1109/AVSS.2009.58}.

\bibitem[Peng et~al.(2025)Peng, Xu, Liu, Wang, and Gao]{peng2025within}
C.~Peng, T.~Xu, D.~Liu, N.~Wang, and X.~Gao.
\newblock Within 3dmm space: Exploring inherent 3d artifact for video forgery detection.
\newblock \emph{IEEE Transactions on Information Forensics and Security}, 2025.

\bibitem[Peng et~al.(2026)Peng, Sun, Chen, Su, Su, and Liu]{peng2026parametric}
C.~Peng, J.~Sun, Y.~Chen, Z.~Su, Z.~Su, and Y.~Liu.
\newblock Parametric gaussian human model: Generalizable prior for efficient and realistic human avatar modeling.
\newblock In \emph{2026 International Conference on 3D Vision (3DV)}, pages 771--782. IEEE, 2026.

\bibitem[Petmezas et~al.(2025)Petmezas, Vanian, Konstantoudakis, Almaloglou, and Zarpalas]{petmezas2025video}
G.~Petmezas, V.~Vanian, K.~Konstantoudakis, E.~E. Almaloglou, and D.~Zarpalas.
\newblock Video deepfake detection using a hybrid cnn-lstm-transformer model for identity verification.
\newblock \emph{Multimedia Tools and Applications}, 84\penalty0 (33):\penalty0 40617--40636, 2025.

\bibitem[Ploumpis et~al.(2019{\natexlab{a}})Ploumpis, Ververas, Sullivan, Moschoglou, Wang, Pears, Smith, Gecer, and Zafeiriou]{ploumpis2020towards}
S.~Ploumpis, E.~Ververas, E.~O. Sullivan, S.~Moschoglou, H.~Wang, N.~E. Pears, W.~Smith, B.~Gecer, and S.~Zafeiriou.
\newblock Towards a complete 3d morphable model of the human head.
\newblock \emph{IEEE transactions on pattern analysis and machine intelligence}, 43\penalty0 (11):\penalty0 4142--4160, 2019{\natexlab{a}}.
\newblock \doi{10.1109/TPAMI.2020.2991150}.

\bibitem[Ploumpis et~al.(2019{\natexlab{b}})Ploumpis, Wang, Pears, Smith, and Zafeiriou]{ploumpis2019combining}
S.~Ploumpis, H.~Wang, N.~Pears, W.~A. Smith, and S.~Zafeiriou.
\newblock Combining 3d morphable models: A large scale face-and-head model.
\newblock In \emph{Computer Vision and Pattern Recognition}, pages 10926--10935. IEEE, 2019{\natexlab{b}}.
\newblock \doi{10.1109/CVPR.2019.01119}.

\bibitem[Ploumpis et~al.(2022)Ploumpis, Moschoglou, Triantafyllou, and Zafeiriou]{ploumpis20223d}
S.~Ploumpis, S.~Moschoglou, V.~Triantafyllou, and S.~Zafeiriou.
\newblock 3d human tongue reconstruction from single" in-the-wild" images.
\newblock In \emph{Computer Vision and Pattern Recognition}, pages 2771--2780, 2022.
\newblock \doi{10.1109/CVPR52688.2022.00279}.

\bibitem[Potamias et~al.(2025)Potamias, Galanakis, Deng, Papaioannou, and Zafeiriou]{potamias2025imhead}
R.~A. Potamias, S.~Galanakis, J.~Deng, A.~Papaioannou, and S.~Zafeiriou.
\newblock Imhead: A large-scale implicit morphable model for localized head modeling.
\newblock In \emph{IEEE International Conference on Computer Vision}, pages 10196--10206. IEEE, 2025.
\newblock \doi{10.1109/ICCV51701.2025.00950}.

\bibitem[Prinzler et~al.(2025)Prinzler, Zakharov, Sklyarova, Kabadayi, and Thies]{giebenhain2025joker}
M.~Prinzler, E.~Zakharov, V.~Sklyarova, B.~Kabadayi, and J.~Thies.
\newblock Joker: Conditional 3d head synthesis with extreme facial expressions.
\newblock In \emph{International Conference on 3D Vision}, 2025.
\newblock \doi{10.1109/3DV66043.2025.00148}.

\bibitem[Qian et~al.(2024)Qian, Kirschstein, Schoneveld, Davoli, Giebenhain, and Nie{\ss}ner]{qian2024gaussianavatars}
S.~Qian, T.~Kirschstein, L.~Schoneveld, D.~Davoli, S.~Giebenhain, and M.~Nie{\ss}ner.
\newblock Gaussianavatars: Photorealistic head avatars with rigged 3d gaussians.
\newblock In \emph{Proceedings of the IEEE/CVF Conference on Computer Vision and Pattern Recognition}, pages 20299--20309, 2024.

\bibitem[Qiu et~al.(2025)Qiu, Zhang, Beeler, Tankovich, Hane, Fanello, Rhemann, and Escolano]{qiu2025chosen}
D.~Qiu, Y.~Zhang, T.~Beeler, V.~Tankovich, C.~Hane, S.~Fanello, C.~Rhemann, and S.~Escolano.
\newblock Chosen: Contrastive hypothesis selection for multi-view depth refinement.
\newblock \emph{Proceedings of the Conference on Robots and Vision}, 2025.
\newblock \doi{10.48550/arXiv.2404.02225}.

\bibitem[Rombach et~al.(2022)Rombach, Blattmann, Lorenz, Esser, and Ommer]{rombach2022high}
R.~Rombach, A.~Blattmann, D.~Lorenz, P.~Esser, and B.~Ommer.
\newblock High-resolution image synthesis with latent diffusion models.
\newblock In \emph{Proceedings of the IEEE/CVF conference on computer vision and pattern recognition}, pages 10684--10695, 2022.

\bibitem[Sohn et~al.(2015)Sohn, Lee, and Yan]{sohn2015learning}
K.~Sohn, H.~Lee, and X.~Yan.
\newblock Learning structured output representation using deep conditional generative models.
\newblock \emph{Advances in neural information processing systems}, 28, 2015.

\bibitem[Sorkine et~al.(2004)Sorkine, Cohen{-}Or, Lipman, Alexa, R{\"{o}}ssl, and Seidel]{Sorkine2004}
O.~Sorkine, D.~Cohen{-}Or, Y.~Lipman, M.~Alexa, C.~R{\"{o}}ssl, and H.~Seidel.
\newblock Laplacian surface editing.
\newblock In J.~Boissonnat and P.~Alliez, editors, \emph{Second Eurographics Symposium on Geometry Processing, Nice, France, July 8-10, 2004}, volume~71 of \emph{{ACM} International Conference Proceeding Series}, pages 175--184. Eurographics Association, 2004.
\newblock \doi{10.2312/SGP/SGP04/179-188}.

\bibitem[Srinivasan et~al.(2021)Srinivasan, Wang, Rojas, Kl{\'a}r, Kavan, and Sifakis]{srinivasan2021learning}
S.~G. Srinivasan, Q.~Wang, J.~Rojas, G.~Kl{\'a}r, L.~Kavan, and E.~Sifakis.
\newblock Learning active quasistatic physics-based models from data.
\newblock \emph{ACM Transactions on Graphics (ToG)}, 40\penalty0 (4):\penalty0 1--14, 2021.

\bibitem[Sun et~al.(2024)Sun, Lv, Ye, Lin, Sheng, Wen, Yu, and Liu]{sun2024diffposetalk}
Z.~Sun, T.~Lv, S.~Ye, M.~Lin, J.~Sheng, Y.-H. Wen, M.~Yu, and Y.-J. Liu.
\newblock Diffposetalk: Speech-driven stylistic 3d facial animation and head pose generation via diffusion models.
\newblock \emph{ACM TOG}, 2024.

\bibitem[Taubin(1995)]{taubin1995}
G.~Taubin.
\newblock A signal processing approach to fair surface design.
\newblock In \emph{SIGGRAPH}, pages 351--358. {ACM}, 1995.

\bibitem[Varol et~al.(2017)Varol, Romero, Martin, Mahmood, Black, Laptev, and Schmid]{varol2017learning}
G.~Varol, J.~Romero, X.~Martin, N.~Mahmood, M.~J. Black, I.~Laptev, and C.~Schmid.
\newblock Learning from synthetic humans.
\newblock In \emph{Proceedings of the IEEE conference on computer vision and pattern recognition}, pages 109--117, 2017.

\bibitem[Wood et~al.(2016)Wood, Baltrušaitis, Morency, Robinson, and Bulling]{wood20163d}
E.~Wood, T.~Baltrušaitis, L.-p. Morency, P.~Robinson, and A.~Bulling.
\newblock A 3d morphable eye region model for gaze estimation.
\newblock In \emph{European conference on computer vision}, pages 297--313. Springer, Springer International Publishing, 2016.
\newblock \doi{10.1007/978-3-319-46448-0_18}.

\bibitem[Wood et~al.(2021)Wood, Baltruvsaitis, Hewitt, Dziadzio, Johnson, Estellers, Cashman, and Shotton]{wood2021fakeit}
E.~Wood, T.~Baltruvsaitis, C.~Hewitt, S.~Dziadzio, M.~Johnson, V.~Estellers, T.~Cashman, and J.~Shotton.
\newblock Fake it till you make it: face analysis in the wild using synthetic data alone.
\newblock In \emph{IEEE International Conference on Computer Vision}, pages 3661--3671. IEEE, 2021.
\newblock \doi{10.1109/ICCV48922.2021.00366}.

\bibitem[Wood et~al.(2022)Wood, Baltrušaitis, Hewitt, Johnson, Shen, Milosavljević, Wilde, Garbin, Sharp, Stojiljković, Cashman, and Julien]{wood20223d}
E.~Wood, T.~Baltrušaitis, C.~Hewitt, M.~Johnson, J.~Shen, N.~Milosavljević, D.~Wilde, S.~J. Garbin, T.~Sharp, I.~Stojiljković, T.~Cashman, and V.~Julien.
\newblock 3d face reconstruction with dense landmarks.
\newblock In \emph{European Conference on Computer Vision}, pages 160--177. Springer, Springer Nature Switzerland, 2022.
\newblock \doi{10.48550/arXiv.2204.02776}.

\bibitem[Wu et~al.(2016)Wu, Bradley, Garrido, Zollhöfer, Theobalt, Gross, and Beeler]{wu2016model}
C.~Wu, D.~Bradley, P.~Garrido, M.~Zollhöfer, C.~Theobalt, M.~Gross, and T.~Beeler.
\newblock Model-based teeth reconstruction.
\newblock \emph{ACM Transactions on Graphics}, 35\penalty0 (6):\penalty0 1--13, 2016.
\newblock \doi{10.1145/2980179.2980233}.

\bibitem[Wu et~al.(2018)Wu, Shiratori, and Sheikh]{wu18incremental}
C.~Wu, T.~Shiratori, and Y.~Sheikh.
\newblock Deep incremental learning for efficient high-fidelity face tracking.
\newblock \emph{ACM TOG}, 2018.

\bibitem[Xu et~al.(2024)Xu, Chen, Li, Zhang, Wang, Zheng, and Liu]{xu2024gaussian}
Y.~Xu, B.~Chen, Z.~Li, H.~Zhang, L.~Wang, Z.~Zheng, and Y.~Liu.
\newblock Gaussian head avatar: Ultra high-fidelity head avatar via dynamic gaussians.
\newblock In \emph{Computer Vision and Pattern Recognition}, pages 1931--1941. IEEE, 2024.
\newblock \doi{10.1109/CVPR52733.2024.00189}.

\bibitem[Xu et~al.(2025)Xu, Su, Wu, and Liu]{xu2025gphm}
Y.~Xu, Z.~Su, Q.~Wu, and Y.~Liu.
\newblock Gphm: Gaussian parametric head model for monocular head avatar reconstruction.
\newblock \emph{IEEE Transactions on Pattern Analysis and Machine Intelligence}, 2025.
\newblock \doi{10.1109/TPAMI.2025.3596331}.

\bibitem[Yan et~al.(2025)Yan, Ward, Wang, Tang, and Du]{yan2025stylemorpheus}
P.~Yan, R.~K. Ward, D.~Wang, Q.~Tang, and S.~Du.
\newblock Stylemorpheus: A style-based 3d-aware morphable face model.
\newblock \emph{arXiv preprint arXiv:2503.11792}, 2025.

\bibitem[Yang et~al.(2022)Yang, Kim, Zoss, G{\"o}zc{\"u}, Gross, and Solenthaler]{yang2022implicit}
L.~Yang, B.~Kim, G.~Zoss, B.~G{\"o}zc{\"u}, M.~Gross, and B.~Solenthaler.
\newblock Implicit neural representation for physics-driven actuated soft bodies.
\newblock \emph{ACM Transactions on Graphics (ToG)}, 41\penalty0 (4):\penalty0 1--10, 2022.

\bibitem[Yang et~al.(2023)Yang, Zoss, Chandran, Gotardo, Gross, Solenthaler, Sifakis, and Bradley]{yang2023implicit}
L.~Yang, G.~Zoss, P.~Chandran, P.~Gotardo, M.~Gross, B.~Solenthaler, E.~Sifakis, and D.~Bradley.
\newblock An implicit physical face model driven by expression and style.
\newblock In \emph{SIGGRAPH Asia 2023 conference papers}, pages 1--12, 2023.

\bibitem[Yang et~al.(2024)Yang, Zoss, Chandran, Gross, Solenthaler, Sifakis, and Bradley]{yang2024learning}
L.~Yang, G.~Zoss, P.~Chandran, M.~Gross, B.~Solenthaler, E.~Sifakis, and D.~Bradley.
\newblock Learning a generalized physical face model from data.
\newblock \emph{ACM Transactions on Graphics (TOG)}, 43\penalty0 (4):\penalty0 1--14, 2024.

\bibitem[Yang et~al.(2019)Yang, Marshak, S{\`y}kora, Ramalingam, and Kavan]{yang2019jaw}
W.~Yang, N.~Marshak, D.~S{\`y}kora, S.~Ramalingam, and L.~Kavan.
\newblock Building anatomically realistic jaw kinematics model from data.
\newblock \emph{The Visual Computer}, 35\penalty0 (6):\penalty0 1105--1118, 2019.

\bibitem[Zhang et~al.(2022)Zhang, Elgharib, Fox, Gu, Theobalt, and Wang]{zhang2022implicit}
C.~Zhang, M.~Elgharib, G.~Fox, M.~Gu, C.~Theobalt, and W.~Wang.
\newblock An implicit parametric morphable dental model.
\newblock \emph{ACM Transactions on Graphics}, 41\penalty0 (6):\penalty0 1--13, 2022.
\newblock \doi{10.1145/3550454.3555469}.

\bibitem[Zhang et~al.(2018)Zhang, Cisse, Dauphin, and Lopez-Paz]{zhang2017mixup}
H.~Zhang, M.~Cisse, Y.~N. Dauphin, and D.~Lopez-Paz.
\newblock mixup: Beyond empirical risk minimization.
\newblock In \emph{International Conference on Learning Representations}, 2018.

\bibitem[Zhang et~al.(2023)Zhang, Rao, and Agrawala]{zhang2023adding}
L.~Zhang, A.~Rao, and M.~Agrawala.
\newblock Adding conditional control to text-to-image diffusion models.
\newblock In \emph{IEEE International Conference on Computer Vision}, pages 3813--3824. IEEE, 2023.
\newblock \doi{10.1109/ICCV51070.2023.00355}.

\bibitem[Zhou and Zafeiriou(2017)]{zhou2017deformable}
Y.~Zhou and S.~Zafeiriou.
\newblock Deformable models of ears in-the-wild for alignment and recognition.
\newblock In \emph{IEEE International Conference on Automatic Face and Gesture Recognition}, pages 626--633. IEEE, IEEE, 2017.
\newblock \doi{10.1109/FG.2017.79}.

\bibitem[Zielonka et~al.(2025)Zielonka, Garbin, Lattas, Kopanas, Gotardo, Beeler, Thies, and Bolkart]{zielonka2025synthetic}
W.~Zielonka, S.~J. Garbin, A.~Lattas, G.~Kopanas, P.~Gotardo, T.~Beeler, J.~Thies, and T.~Bolkart.
\newblock Synthetic prior for few-shot drivable head avatar inversion.
\newblock In \emph{Computer Vision and Pattern Recognition}, pages 10735--10746. IEEE, 2025.
\newblock \doi{10.1109/CVPR52734.2025.01003}.

\bibitem[Zoss et~al.(2018)Zoss, Bradley, Bérard, and Beeler]{zoss2018empirical}
G.~Zoss, D.~Bradley, P.~Bérard, and T.~Beeler.
\newblock An empirical rig for jaw animation.
\newblock In \emph{ACM Transactions on Graphics}, pages 1--12. Association for Computing Machinery (ACM), 2018.

\bibitem[Zoss et~al.(2019)Zoss, Beeler, Gross, and Bradley]{zoss2019accurate}
G.~Zoss, T.~Beeler, M.~Gross, and D.~Bradley.
\newblock Accurate markerless jaw tracking for facial performance capture.
\newblock \emph{ACM Transactions on Graphics (TOG)}, 38\penalty0 (4):\penalty0 1--8, 2019.

\end{thebibliography}

\end{document}